%% file: main.tex
\documentclass[10pt,twocolumn,letterpaper]{article}

\usepackage{cvpr}              %
\usepackage[utf8]{inputenc} %
\usepackage[T1]{fontenc}    %
\usepackage{url}            %
\usepackage{booktabs}       %
\usepackage{amsfonts}       %
\usepackage{nicefrac}       %
\usepackage{microtype}      %
\usepackage{xcolor}         %
\usepackage[symbol]{footmisc}

\usepackage[export]{adjustbox}
\usepackage{tikz}
\usetikzlibrary{shapes.geometric, calc}
\usepackage{wrapfig}

\usepackage{amssymb}
\usepackage{pifont}
\usepackage{graphicx}
\graphicspath{{images/}}
\usepackage{natbib}
\usepackage{algorithm2e}

\usepackage{diagbox}
\usepackage{amsmath}
\usepackage{amssymb}
\usepackage{mathtools}
\usepackage{amsthm}
\usepackage{siunitx}
\usepackage{caption}
\usepackage{capt-of}
\definecolor{cvprblue}{rgb}{0.21,0.49,0.74}
\usepackage[pagebackref,breaklinks,colorlinks,allcolors=cvprblue]{hyperref}
\usepackage[capitalize,noabbrev]{cleveref}
\usepackage[skip=2pt,subrefformat=parens]{subcaption}
\usepackage{tabularx}
\usepackage{colortbl}

\usepackage{colortbl}
\usepackage{booktabs}
\usepackage{listings}
\usepackage{xcolor}
\usepackage{microtype}
\theoremstyle{plain}

\theoremstyle{definition}

\theoremstyle{remark}

\usepackage{import}
\usepackage{lipsum}
\usepackage{multirow}
\usepackage{xspace}
\usepackage{afterpage}
\usepackage{fancyhdr}
\usepackage{setspace}
\usepackage{colortbl} %

\DeclareMathOperator*{\softmax}{softmax}
\DeclareMathOperator*{\argmax}{arg\,max}

\crefname{equation}{Eq.}{Eqs.}

\input{preamble}

\def\paperID{12908} %
\def\confName{CVPR}
\def\confYear{2025}

\title{
GLASS: Guided Latent Slot Diffusion for Object-Centric Learning
}
\author{Krishnakant Singh$^{1}$
\quad\quad
Simone Schaub-Meyer$^{1, 2}$ 
\quad\quad
Stefan Roth$^{1, 2}$ \\
$^1$ Department of Computer Science, TU Darmstadt \;\; $^2$ hessian.AI \\
}
\fancypagestyle{firststyle}
{

	\fancyhf{}
	\lfoot{{\footnotesize\begin{spacing}{.5}\parbox{\linewidth}{\vspace{2.5em}%
					To appear in \emph{Proceedings of the IEEE/CVF Conference on Computer Vision and Pattern Recognition}, Nashville, TN, USA, 2025.%
					\\\hrule\vspace{\baselineskip}
					\copyright~2025 IEEE. Personal use of this material is permitted. Permission from IEEE must be obtained for all other uses, in any current or future media, including reprinting/republishing this material for advertising or promotional purposes, creating new collective works, for resale or redistribution to servers or lists, or reuse of any copyrighted component of this work in other works. 
	}\end{spacing}}}
}
\begin{document}
\maketitle
\input{sec/abstract}    
\input{sec/introduction}
\input{sec/related_work}

\input{sec/methodology}
\input{sec/experiments}

\input{sec/conclusion}

{
    \small
    \bibliographystyle{ieeenat_fullname}
    \bibliography{main}
}

 \input{suppl}

\end{document}

%% file: preamble.tex
\newcommand{\slot}{\mathbf{S}}
\newcommand{\attn}{\mathbf{A}}
\newcommand{\feat}{\mathbf{H}}
\newcommand{\inp}{\mathbf{I}_{\text{inp}}}
\newcommand{\genimg}{\mathbf{I}_{\text{gen}}}
\newcommand{\reconimg}{\mathbf{I}_{\text{recon}}}

\newcommand{\genfeat}{\mathbf{F}_{\text{inp}}}
\newcommand{\reconfeat}{\mathbf{F}_{\text{recon}}}

\newcommand{\maskprev}{\mathbf{M}_{\text{ref}}}
\newcommand{\mask}{\mathbf{M}_{\text{gen}}}
\newcommand{\mycaption}{\mathcal{P}_{\text{cap}}}
\newcommand{\myprompt}{\mathcal{P}}
\newcommand{\myclass}{\mathcal{C}}
\newcommand{\condtext}{\mathbf{Y}}

\newcommand{\rot}[2][l]{\rotatebox[origin=#1]{90}{#2}}  %

\newcommand\cre[1]{#1}

\DeclareRobustCommand{\circled}[2]{\tikz[baseline=(char.base)]{
\node[shape=circle, fill=#2, draw=black, inner sep=0.40pt] (char) {#1};}}
\definecolor{turquoise}{cmyk}{0.65,0,0.1,0.3}
\definecolor{purple}{rgb}{0.65,0,0.65}
\definecolor{dark_green}{rgb}{0, 0.5, 0}
\definecolor{orange}{rgb}{0.8, 0.6, 0.2}
\definecolor{dark_orange}{rgb}{0.7, 0.6, 0.3}
\definecolor{darkred}{rgb}{0.6, 0.1, 0.05}
\definecolor{blueish}{rgb}{0.0, 0.3, .6}
\definecolor{light_gray}{rgb}{0.7, 0.7, .7}
\definecolor{pink}{rgb}{1, 0, 1}
\definecolor{cyan}{rgb}{0., 1, 1}

\definecolor{myblue}{RGB}{115, 191, 249}
\definecolor{mypink}{RGB}{241, 154, 200}
\definecolor{myyellow}{RGB}{249, 219, 88}
\definecolor{myturquoise}{RGB}{152, 249, 234}
\definecolor{mysalmon}{RGB}{237, 110, 88}
\definecolor{mygreen}{RGB}{164, 247, 105}
\definecolor{myblueish}{RGB}{213, 228, 247}
\definecolor{checkyes}{rgb}{1.0, 1.0, 1.0}
\definecolor{crossno}{rgb}{1.0, 1.0, 1.0}

\definecolor{best}{HTML}{d7191c}
\definecolor{secondbest}{HTML}{0571b0}

\newcommand{\best}[1]{\textbf{#1}}
\newcommand{\secondbest}[1]{\underline{#1}}

\newcommand{\cmark}{\text{\ding{51}}}%
\newcommand{\xmark}{\text{\ding{55}}}%
\def \y {$\cmark$\cellcolor{checkyes}}
\def \n {$\xmark$\cellcolor{crossno}}

\definecolor{citecolor}{RGB}{0, 113, 188}

\Crefname{section}{Section}{Sections}
\Crefname{table}{Table}{Tables}

\crefname{figure}{Fig.}{Figs.}
\makeatletter
\DeclareRobustCommand\onedot{\futurelet\@let@token\@onedot}
\def\@onedot{\ifx\@let@token.\else.\null\fi\xspace}
\newcommand{\ourmethod}{GLASS\xspace}
\newcommand{\ourmethodDagger}{GLASS$^\dagger$\xspace}

\newcommand*{\myparagraph}[1]{\smallskip\noindent\textbf{#1}\hspace{0.5em}}
\def\eg{\emph{e.g}\onedot} 
\def\ie{\emph{i.e}\onedot} 
\def\cf{\emph{cf}\onedot}

\renewcommand{\paragraph}[1]{\vspace{1.25mm}\noindent\textbf{#1}}

\newcolumntype{x}[1]{>{\centering\arraybackslash}p{#1pt}}
\newcolumntype{y}[1]{>{\raggedright\arraybackslash}p{#1pt}}
\newcolumntype{z}[1]{>{\raggedleft\arraybackslash}p{#1pt}}

\newcommand{\changed}[1]{\textcolor{black}{#1}}

\usepackage{siunitx} %

%% file: sec/abstract.tex
\begin{abstract}
\label{sec:abs}
Object-centric learning aims to decompose an input image into a set of meaningful object files (slots). 
These latent object representations enable a variety of downstream tasks. 
Yet, object-centric learning struggles on real-world datasets, which contain multiple objects of complex textures and shapes in natural everyday scenes. 
To address this, we introduce \textbf{G}uided \textbf{La}tent \textbf{S}lot Diffu\textbf{s}ion (GLASS), a novel slot-attention model that learns in the space of generated images and uses semantic and instance guidance modules to learn better slot embeddings for various downstream tasks.
Our experiments show that \ourmethod surpasses state-of-the-art slot-attention methods by a wide margin on tasks such as (zero-shot) object discovery and conditional image generation for real-world scenes. %
Moreover, \ourmethod enables the first application of slot attention to the compositional generation of complex, realistic scenes.
\footnote{Project page and code: \href{https://visinf.github.io/glass/}{https://visinf.github.io/glass/}}
\end{abstract}

%% file: sec/introduction.tex
\section{Introduction} \label{sec:intro}
    Humans perceive a scene as a collection of objects \cite{kahneman1992reviewing}. 
    Such a decomposition of the scene into objects makes humans capable of higher cognitive tasks like control, reasoning, and the ability to generalize to unseen experiences \cite{greff2020binding}. 
    Building on these ideas, object-centric learning (OCL) aims to decompose a scene into compositional and modular symbolic components. 
    OCL methods bind these components to latent (neural) representations, enabling such models to be applied to tasks like causal inference \cite{scholkopf2021causal}, reasoning \cite{assouel2022object}, control \cite{biza2022binding}, and out-of-distribution generalization \cite{dittadi2021generalization}.

\input{tables/nips_tabs/teaser_table}

    Slot-attention models \cite{locatello2020object}, a popular class of OCL methods, decompose an image into a set of latent representations where each element, called slot, competes to represent a certain part of the image. 
    Slot-attention methods can be categorized as form of representation learning, where the representation (slots) facilitates various downstream tasks such as property prediction~\cite{dittadi2021generalization}, image reconstruction~\cite{jiang2023object}, image editing~\cite{wu2023slotdiffusion}, and object discovery~\cite{seitzer2022bridging}. However, numerous promising slot-attention methods \cite{locatello2020object,singh2021illiterate,singh2022simple} have remained limited to synthetic and simple datasets \cite{greff2022kubric,johnson2017clevr,karazija2021clevrtex,elich2022weakly}.
    Some recent methods \cite{seitzer2022bridging,kakogeorgiou2024spot,jiang2023object,wu2023slotdiffusion} use powerful, modern encoder  \cite{oquab2023dinov2,caron2021emerging} and decoder networks \cite{rombach2022high} to scale to complex real-world imagery \cite{lin2014microsoft,everingham2010pascal}. Yet, these models remain restricted to object discovery, lacking the versatility to reconstruct or perform compositional generation of realistic images. Moreover, the quality of the obtained slot representations remains limited as witnessed by both qualitative and quantitative results, which show the slots to suffer from the issue of over-segmentation (segmenting an object into multiple slots), under-segmentation (segmenting multiple objects into one slot), or imprecise object boundaries. This over- and under-segmentation issue is also known as the part-whole hierarchy ambiguity \cite{hinton1979some,hinton1990mapping,hinton2023represent}.
    
    To overcome the above issues, we propose \textbf{G}uided \textbf{La}tent \textbf{S}lot Diffu\textbf{s}ion (GLASS), a \cre{weakly-supervised} slot attention-based model that uses a pre-trained diffusion decoder for reconstructing the input image and an MLP decoder for reconstructing the encoder features. GLASS relies on two key observations: 
    \emph{(1)} Learning in the space of images generated using diffusion models allows to generalize well to real images because the distribution of the generated images closely mimics the real data distribution \cite{fan2023scaling,sariyildiz2023fake,tian2024stablerep,singh2024is}; and 
    \emph{(2)} learning with generated images allows us to use a pre-trained diffusion model, such as Stable Diffusion \cite{rombach2022high}, as a pseudo ground-truth generation engine.
    To this end, \ourmethod relies on a semantic guidance module, which uses the diffusion decoder to generate the pseudo-semantic mask. The semantic guidance module helps \ourmethod address over-segmentation issues and obtain precise boundaries. 
 
    However, using semantic guidance alone biases the slots to semantic classes instead of instances in an image, causing under-segmentation. To resolve this issue, we use an instance module in the form of an MLP decoder \cre{similar to \cite{seitzer2022bridging}, which reconstructs the encoder features to counteract slots drifting towards semantic classes}. %
    This enables the slots to learn better slot embeddings, which are more instance centric.
    \cre{While some of \ourmethod{}'s modules have been considered in isolation, its novel combination of semantic and instance guidance modules coupled with a diffusion decoder enable} it to faithfully reconstruct / conditionally generate the input image. More importantly, \ourmethod for the first time enables the compositional generation of complex real-world scenes with slot-attention methods. \cref{fig:intro:teaser} illustrates the high-level architecture and the downstream tasks our model supports.
   
    Through our experiments, we show that \ourmethod outperforms existing \cre{state-of-the-art (SotA)} OCL methods \cite{seitzer2022bridging,kakogeorgiou2024spot,jiang2023object,wu2023slotdiffusion}, significantly improving instance-level object discovery (ca.\ +9\,\% mIoU$_i$ on VOC \cite{everingham2010pascal} and +5\,\% mIoU$_i$ on COCO \cite{lin2014microsoft}). Our method also outperforms SotA OCL methods for (zero-shot) instance-level segmentation (on Object365 \cite{shao2019objects365} and CLEVRTex \cite{karazija2021clevrtex}). 
    \ourmethod further establishes a new SotA FID score among OCL methods for conditional image generation. \cre{\ourmethod also makes compositional generation possible with slot attention for complex real-world scenes}. 
    Moreover, we find that our approach surpasses language-based methods \cite{wu2023diffumask,luo2023segclip,ranasinghe2023perceptual,zhou2022extract} for semantic-level object discovery.
    Finally, we show that \ourmethod outperforms \cre{other} weakly-supervised SotA OCL models \cite{jiang2023object} that rely on extra information like bounding boxes or knowing the number of objects in a scene. 

%% file: tables/nips_tabs/teaser_table.tex
\begin{figure}[t]
    \centering    
    \includegraphics[width=0.98\linewidth]{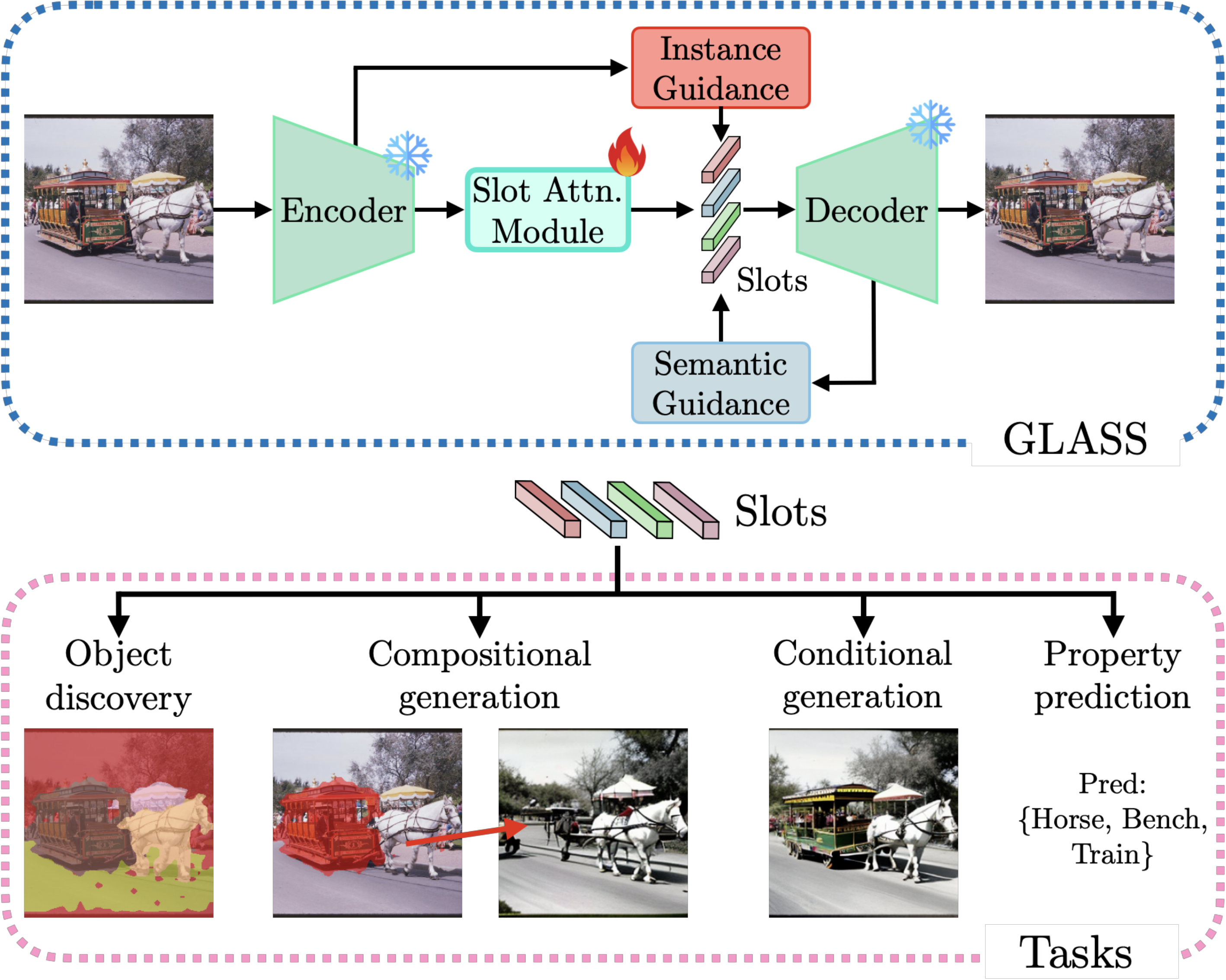}
    \vspace{-0.5em}
    \caption{\emph{(top)} \textbf{High-level architecture of \ourmethod.} \ourmethod employs semantic and instance guidance modules to generate a semantic guidance signal using the decoder network and an instance guidance signal using the encoder features. This helps our method to learn superior slot embeddings for various downstream tasks compared to existing slot-attention methods. 
    \emph{(bottom)} \textbf{\ourmethod can perform multiple tasks} using the learned slot embeddings, such as object discovery, compositional generation, conditional generation (reconstruction), and instance-level property prediction. %
    }
    \vspace{-0.5em}
    \label{fig:intro:teaser}
\end{figure}

%% file: sec/related_work.tex
\section{Related Work}\label{sec:rel_work}
    \input{tables/nips_tabs/baseline_cmp}    
    \textbf{Object-centric learning} decomposes a multi-object scene into a set of composable and meaningful entities using an autoencoding objective \cite{burgess2019monet,crawford2019spatially,engelcke2020genesis,eslami2016attend,greff2019multi,greff2020binding,kabra2021simone,locatello2020object,singh2022simple}. 
    OCL methods are object-level representation learning approaches that can be employed for various downstream tasks (\cf \cref{tab:rel_work:baselines}).
    Among OCL approaches, slot-attention methods have proven the most effective; they employ an architectural inductive bias to learn object embeddings, so-called ``slots'', from the input image. 
    Until recently, a major obstacle for slot attention had been its poor performance on real-world images \cite{yang2022promising}. 
    This was partially alleviated using large-scale pre-trained models as encoder \cite{seitzer2022bridging} and decoder \cite{wu2023slotdiffusion,jiang2023object}, which allowed to apply slot attention beyond synthetic imagery.
    Yet, these models still suffer from the part-whole hierarchy ambiguity, hampering the quality of the learned slot embedding, resulting in poor downstream performance. 
    Our method aims to alleviate this issue using our proposed semantic and instance guidance modules.

    \myparagraph{Weakly-supervised object-centric learning.}
    Several works have tried to tackle the part-whole hierarchy ambiguity plaguing OCL with additional weak supervision signals. Video-based OCL methods used motion \cite{kipf2022conditional,tangemann2023unsupervised} and depth cues \cite{elsayed2022savi}, while image-based OCL methods have used position \cite{kim2023shatter} and shape \cite{elich2022weakly} information.
    Existing weakly-supervised image-based OCL methods \cite{kim2023shatter,elich2022weakly} remain limited to synthetic datasets, while we focus on complex real-world scenes. 
    \ourmethod also uses auxiliary information, \cre{here} in the form of \emph{automatically} generated captions. To show the effectiveness of our method, we additionally compare \ourmethod to a weakly-supervised variant of StableLSD \cite{jiang2023object} (since it is closest in capabilities to \ourmethod, see \cref{tab:rel_work:baselines}).

    \myparagraph{Semantic-level object discovery.} 
    Recently, there has been a large interest in using pre-trained features from large-scale foundational models \cite{oquab2023dinov2,caron2021emerging, rombach2022high, radford2021learning} for semantic segmentation. Some of these models \cite{nguyen2023dataset,khani2023slime,wu2023diffumask,karazija2024diffusion,pnvr2023ld,xu2023open,ding2023open,ranasinghe2023perceptual,mukhoti2023open,cha2023learning} rely on language cues like image-level labels or captions to extract features, which are suitable for semantic segmentation. 
    Other methods like \cite{zadaianchuk2023unsupervised,melas2022deep,namekata2024emerdiff,couairon2024zeroshot} do not require any additional information and use clustering or graph cuts with pre-trained features for semantic segmentation.
    Unlike OCL methods, these approaches are specifically designed to perform semantic-level segmentation, \ie, they cannot distinguish between objects of the same class. Also, they cannot generate images conditionally or compositionally, nor perform object-level reasoning (see \cref{tab:rel_work:baselines}). We compare such methods with a semantic-focused version of \ourmethod to show its efficacy on semantic-level object discovery.

%% file: tables/nips_tabs/baseline_cmp.tex
\begin{table}
\smallskip
\centering
\scriptsize
\begin{tabularx}{\columnwidth}{@{}l*{6}X|*{5}X@{}}
\toprule
 & \multicolumn{6}{c}{Semantic-level OD methods} 
 & \multicolumn{5}{c}{OCL methods}
 
 \\
 \cmidrule(lr){2-7}
 \cmidrule(lr){8-12}
 & \rot{DeepSpectral \cite{cho2021picie}} 
 & \rot{SegCLIP \cite{luo2023segclip}} 
 & \rot{CLIPpy \cite{ranasinghe2023perceptual}}
 & \rot{DiffuMask \cite{wu2023diffumask}}
 & \rot{Dataset Diff. \cite{nguyen2023dataset}}
 & \rot{EmerDiff \cite{namekata2024emerdiff}}
 & \rot{Slot Attention \cite{locatello2020object}}
 & \rot{DINOSAUR \cite{seitzer2022bridging}}
 & \rot{SPOT \cite{kakogeorgiou2024spot}}
 & \rot{StableLSD \cite{jiang2023object}}
 & \rot{GLASS (ours)} 
 \\ 
\midrule
{\emph{(1)} iOD} & \n & \n & \n & \n & \n & \n & (\y) & \y & \y & \y & \y\\
{\emph{(2)} sOD} & \y & \y & \y & \y & \y & \y & (\y) & \y & \y & \y & \y\\
{\emph{(3)} Latent object file (PP)} & \n & \n & \n & \n & \n & \n & (\y) & \y & \y & \y & \y \\
{\emph{(4)} Cond.\ Gen.\ (CG)}  & \n & \n & \n & \n & \n & \n & (\y) & \n & \n & \y & \y \\
{\emph{(5)} Comp.\ Gen.\ (CPG) } & \n & \n & \n & \n & \n & \n & (\y) & \n & \n & (\y) & \y \\
\bottomrule
\end{tabularx}
\vspace{-0.5em}
\caption{
    \textbf{\ourmethod{}'s capabilities compared with prior work for solving downstream tasks on \emph{real-world scenes}.}
    The rows indicate if each method
    \emph{(1)} can perform instance-level object discovery (OD);
    \emph{(2)} can perform semantic-level OD;
    \emph{(3)} provide latents for each object, which enables instance-level property prediction;
    \emph{(4)} can reconstruct the given image from its latents; and 
    \emph{(5)} can compositionally generate new scenes.
    (\y): limited performance.
}
\vspace{-0.5em}
\label{tab:rel_work:baselines}
\end{table}

%% file: sec/methodology.tex
\input{figures/main_network_updated}
\section{Preliminaries} \label{sec:prelims}
\textbf{Slot attention} \cite{locatello2020object} is an iterative refinement scheme based on a set $\slot \in \mathbb{R}^{O \times d_{\text{slots}}}$, composed of $O$ slots of dimension $d_\text{slots}$, which are initialized either randomly or via a learned function. 
Once initialized, the slot representations are updated iteratively using a GRU network \cite{cho2014learning} based on the feature matrix $\feat \in \mathbb{R}^{N\times d_{\text{input}}}$ of the encoded input image, containing $N$ feature vectors of dimension $d_\text{input}$, and the previous state of the slots. 
Slot attention uses standard dot-product attention~\cite{vaswani2017attention} for computing the attention matrix $\attn \in \mathbb{R}^{N \times O}$, \emph{normalized across slots}. This normalization causes the slots to compete with each other, leading to a meaningful decomposition of the input image. The slots are updated using a weighted combination of the input features $\feat$ and the computed attention matrix $\attn$. 
Formally, we have %
\begin{equation}\label{eq:slot_attn}
\begin{split}
    \hat{\slot} &= \left(\frac{\attn_{i, j}}{\sum_{l=1}^{N}\attn_{l, j}}\right)_{i,j}^\top v(\feat) \\ 
    &\text{with}\ \attn(\slot, \feat) = \softmax \left(\frac{k(\feat) q(\slot)^\top}{\sqrt{D}}\right), 
\end{split}
\end{equation}
where $k$, $q$, and $v$ are learnable linear functions for mapping the slots and input features to the same $D$ dimensions. The updated set of slots $\hat{\slot}$ is fed into a decoder model to reconstruct the input. 
The decoder model can be a simple MLP \cite{watters2019spatial}, a transformer \cite{singh2022simple}, or a diffusion model \cite{wu2023slotdiffusion, jiang2023object}.
Slot-attention methods are trained using the mean squared error loss between the input and reconstructed input signal. 

\smallskip\noindent\textbf{Latent diffusion models (LDM)} \cite{rombach2022high} learn to generate an image by first iteratively destructing the image by adding Gaussian noise at each time step. 
This noising process is called the ``forward process''. 
The ``reverse process'', or generation step, then involves learning a neural network $\epsilon_{\theta}$ that predicts the noise added in each forward diffusion step and removes the noise from the noisy image \changed{at each time step. An additional conditioning signal, most commonly in the form of text, is provided to the diffusion model for enabling conditional generation from the diffusion model.}
The parameters $\theta$ of $\epsilon_{\theta}$ are learned by minimizing the mean squared error between the predicted and ground-truth noise added for each time step in the denoising process. Once trained, an image can be generated by sampling a random noise vector and running the reverse process with a given conditioning signal.
The most common choice for the denoising network $\epsilon_{\theta}$ is a U-Net \cite{ronneberger2015u} with layers of self- and cross-attention at multiple resolutions. The cross-attention layers cross-attend between the conditioning signal and the pixel features. 

\section{Guided Latent Slot Diffusion (GLASS)}\label{sec:gsa}
\ourmethod is based on training a slot-attention module on the features of a DINOv2 \cite{oquab2023dinov2} (encoder) model and uses a pre-trained Stable Diffusion (SD) model \cite{rombach2022high} (decoder) to reconstruct the image, as well as a small MLP model to reconstruct the encoder features. \ourmethod leverages the diffusion decoder and a pre-trained caption generation model \cite{li2023blip} to create a guidance signal (segmentation masks) to guide slots. 

A key design choice in our proposed method is to learn the slot-attention module in the space of generated images from a pre-trained diffusion model. 
This enables us to use the cross-attention layers in the U-Net \cite{ronneberger2015u} of the diffusion decoder for obtaining the semantic mask for a given image. 
Let us now describe each step in detail. 

\myparagraph{Conditional image generation.}
Given an input image $\inp$, we first pass it through a caption generator (BLIP\nobreakdash-2~\cite{li2023blip}) %
to generate a caption $\mycaption$ that describes the input image. 
We extract the nouns from the generated caption using a part-of-speech (POS) tagger \cite{bird2004nltk} and retain those nouns, $\myclass = \{\text{c}_1, \text{c}_2, \ldots, \text{c}_k\}$, that belong to the set of class labels of the COCO dataset \cite{lin2014microsoft}. 
We then create a prompt, $\myprompt=[\mycaption; \myclass]$, by concatenating the generated caption and the extracted class labels. 
This prompt $\myprompt$ is fed into a text embedder, here  CLIP \cite{radford2021learning}, to obtain an embedding $\condtext \in \mathbb{R}^{U\times d_\text{token}}$, where $U$ is the number of tokens of dimension $d_\text{token}$. 
We then generate an image $\genimg$ by sampling random noise %
from $\mathcal{N}(0, I)$ and running the ``reverse process'' on a pre-trained diffusion model with $\condtext$ as a conditioning signal.

\myparagraph{Pseudo ground-truth generation module.} For extracting the cross-attention map at time $t$ for layer $l$ from the diffusion model,  we create a new prompt consisting of a single token, namely one of the class tokens from $\myclass$. The cross-attention map for the target label can be computed using standard dot-product attention
between the linear projections of the target class-label embedding and the noisy image features %
into a common space.
This is done for each target class label in $\myclass$. 
The final cross-attention map $\mathbf{A}_{\text{CA}} \in [0, 1]^{H \times W \times \cre{k}} $ is obtained by resizing and averaging the extracted cross-attention maps across different time steps and resolutions. 
Here, $H$ and $W$ are the sizes of the input embedding 
and \cre{$k=|\mathcal{C}|$} is the number of target classes. 
The obtained cross-attention maps are often noisy and require further refinement. 
Recently, several works have addressed the problem of refining such cross-attention maps \cite{khani2023slime,nguyen2023dataset,wu2023diffumask}. We follow \cite{nguyen2023dataset} and use self-attention maps for refining the cross-attention maps. 
In particular, the refined mask $\maskprev$ is obtained by exponentiating the self-attention map $\mathbf{A}_{\text{SA}} \in [0, 1]^{H \times W \times H \times W} $ and multiplying with the cross-attention map $\mathbf{A_{\text{CA}}}$ as described in \cite{nguyen2023dataset}.
The final semantic mask $\mask$ is obtained by taking the pixel-wise $\argmax$ of $\maskprev$ for all target class labels in $\myclass$ to find which class is responsible for a given pixel. Finally, a range-based thresholding is used to classify each pixel as foreground or background.  See supplement for details. \cre{Note that while our method for pseudo ground-truth generation is akin to \cite{nguyen2023dataset}, \ourmethod combines it within a Hungarian matching module and guiding slots, going beyond prior work.}

\myparagraph{Slot matching.} Once the images $\genimg$ and their corresponding pseudo ground-truth semantic masks $\mask$ are generated, we can use these semantic masks to guide the slots.  
First, we pass the generated image $\genimg$ through the encoder and the slot-attention module to obtain a slot decomposition. 
We extract the predicted masks for each slot using the attention matrix $\attn(\slot, \feat)$ from \cref{eq:slot_attn} and resize them to the resolution of the generated semantic mask $\mask$. 
We then assign each predicted mask to the components of the generated semantic masks. 
This is akin to solving a bipartite matching problem for which we use Hungarian matching \cite{kuhn1955hungarian}. Formally, given $O$ slots with their predicted masks and a semantic mask containing $F$ segments, the binary matching matrix $\mathbf{P}\in\{0,1\}^{O\times F}$ can be computed using the Hungarian algorithm that minimizes the cost $c_{i, j}$ of assigning slot $i$ to segment $j$ in the generated mask $\mask$:
\begin{equation}
    \arg\min_{\mathbf{P}} \sum_{i=1}^{O} \sum_{j=1}^{F} -c_{i, j} p_{i, j}, 
\end{equation}
where $p_{i, j} \in \{0,1\}$ indicates whether slot $i$ is matched with segment $j$ of $\mask$.
The optimization is constrained to assign each slot to one and only one segment. 
The cost $c_{i, j}$ is given by the mean Intersection over Union (IoU) between the predicted mask of slot $i$ and segment $j$ of the generated semantic mask $\mask$.

\myparagraph{Loss function.}
Once the assignment is complete, our guided slot-attention model is trained end-to-end using the mean squared error loss ($\mathcal{L}_{\text{MSE}}$) between the generated image $\genimg$ and reconstructed image $\reconimg$, as well as our (semantic) \emph{guidance loss}, \ie, a binary cross-entropy loss ($\mathcal{L}_{\text{BCE}}$) between $\mask$ and the predicted mask from the slots $\attn(\slot, \feat)$.
The binary cross-entropy loss is only computed on the matched slots, according to the matching matrix $\mathbf{P}\equiv\mathbf{P}(\mask, \attn(\slot, \feat))$. 
However, simply using the image reconstruction and semantic guidance loss would lead the slot representation to drift towards semantic classes and not to objects. \cre{Inspired by \cite{seitzer2022bridging}, which utilized a feature reconstruction loss for learning better slot representations, we utilize feature reconstruction to tackle the semantic drift.}
This \emph{instance guidance loss} is given by the mean squared error between the input ($\genfeat$) and reconstructed ($\reconfeat$) features (see \cref{fig:method:network}). Our full loss is given by
\begin{equation}
\begin{split}
    \mathcal{L} = \underbrace{\mathcal{L}_{\text{MSE}}(\genimg, \reconimg)}_{\color{teal}\text{Recon. Loss}\ (\mathcal{L}_{\text{Recon}})} 
    & + \lambda_{s} \underbrace{\mathcal{L}_{\text{BCE}}(\mathbf{P}(\mask, \attn(\slot, \feat)))}_{ \color{blue}\text{Semantic Guidance} \ (\mathcal{L}_{\text{semantic}})} \\
    &+ \lambda_{i} \underbrace{\mathcal{L}_{\text{MSE}}(\genfeat, \reconfeat)}_{ \color{red}\text{Instance Guidance}\ (\mathcal{L}_{\text{instance}})}.\raisetag{28pt}
\end{split}
\end{equation}

The semantic guidance loss helps learn a slot representation that adheres to object boundaries and does not split the object into multiple slots (\ie, avoids over-segmentation) but causes the slots to focus on semantics and not on instances. The feature reconstruction loss helps with the semantic drift problem as features from a pre-trained ViT model already exhibit instance-aware properties \cite{engstler2023understanding}, but using them without semantic guidance results in over- and under-segmentation issues. Thus, when instance and semantic guidance are coupled, the slots are bound to the instances instead of semantics and alleviate the part-whole ambiguity, \cf also \cref{fig:ablation:loss}.

We separate the training process of \ourmethod into two phases: In phase 1, only the slot-attention module and MLP decoder are trained. This helps in learning slot embeddings that bind to instances. In phase 2, we jointly train both the slot-attention module with diffusion and MLP decoders. In this phase, we use a very small learning rate for the slot attention and MLP decoder modules (essentially frozen) and a higher learning rate for the diffusion decoder. The second phase helps the diffusion decoder align to slot embeddings and produce high-fidelity images. Unless otherwise stated, we use $\lambda_{s}=0.7$ and $\lambda_{i}=0.9$ for all our experiments.
\Cref{fig:method:network} shows our full architecture and illustrates each step.
Further details are provided in the supplement. 

%% file: figures/main_network_updated.tex
\begin{figure*}[t] 
    \centering
    \includegraphics[width=0.75\textwidth]{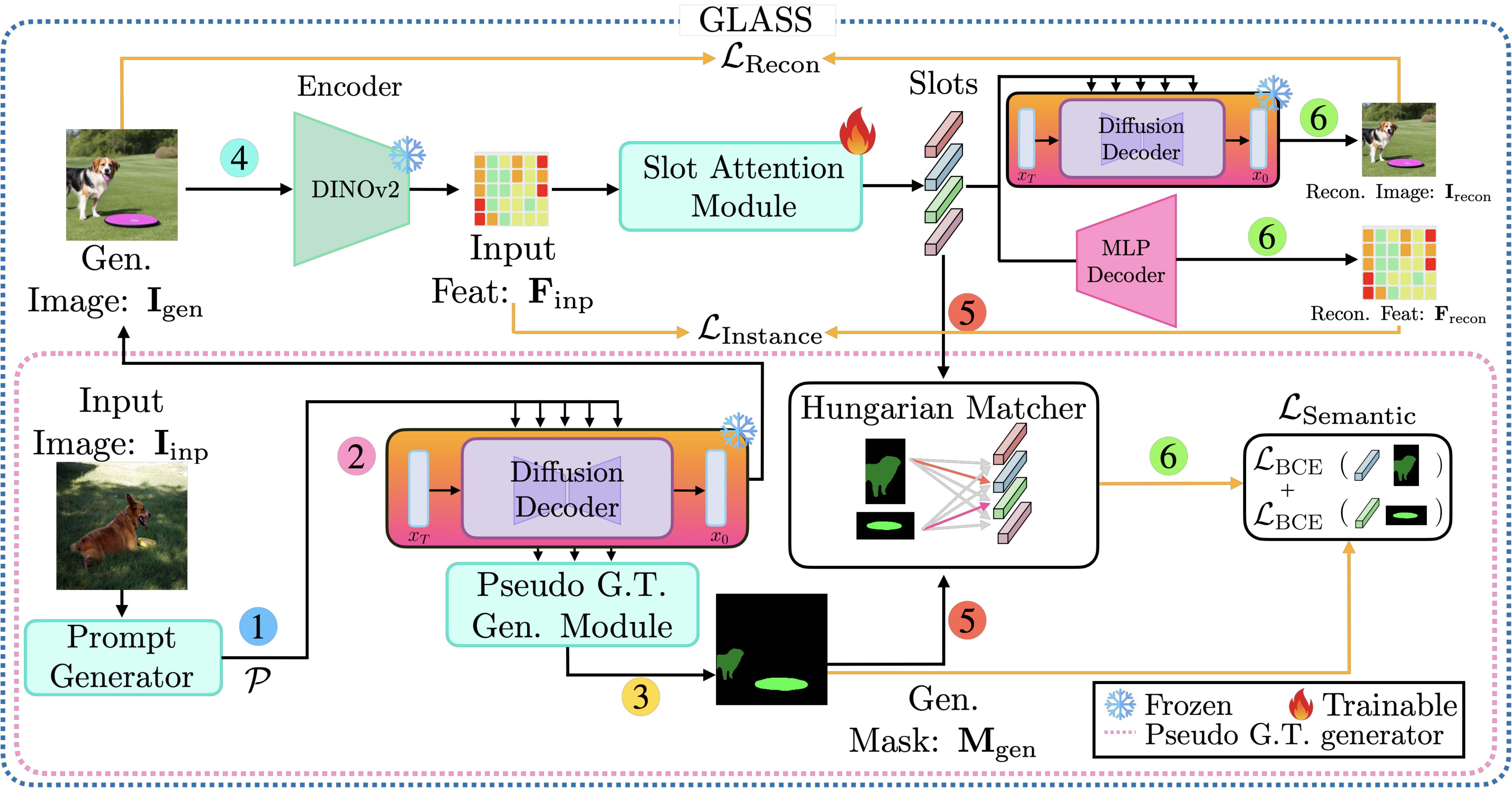}
    \vspace{-0.5em}
    \caption{ \textbf{Network architecture of GLASS.}
    \circled{1}{myblue} The input image $\inp$ is fed to a prompt generator for generating a prompt $\myprompt$, obtained by concatenating the generated caption $\mycaption$ and the extracted class labels from $\mycaption$. \circled{2}{mypink} A random noise vector, along with the generated prompt $\myprompt$, is used to generate an image $\genimg$ using a pre-trained diffusion decoder module. \circled{3}{myyellow} The cross-attention layers of the diffusion model, along with self-attention layers, are used in the pseudo ground-truth generation module to generate the semantic mask $\mask$ for $\genimg$.  \circled{4}{myturquoise} The generated image is passed through an encoder model (DINOv2) followed by a slot-attention module to generate slots.  \circled{5}{mysalmon} The slots are matched with their corresponding object masks from $\mask$ using the Hungarian matcher module. \circled{6}{mygreen} The slot-attention module is trained end-to-end using the mean squared error ($\mathcal{L_{\text{Recon}}}$) between the reconstructed ($\reconimg$) and the generated ($\genimg$) image, and our \emph{semantic} ($\mathcal{L}_{\text{Semantic}}$) and \emph{instance} ($\mathcal{L}_{\text{Instance}}$) guidance losses. \ourmethod is trained on generated images only; the real image is used for prompt generation.
    }
    \label{fig:method:network}
    \vspace{-0.5em}
\end{figure*}

%% file: sec/experiments.tex
\section{Experiments}\label{sec:exp}
Our main goal is to learn better representations of objects, \ie, slot embeddings. To assess the effectiveness of the learned representation, we evaluate on various tasks such as object discovery, instance-level property prediction, reconstruction, and compositional generation.
\ourmethod uses generated captions from BLIP-2 \cite{li2023blip}, which is trained on image-caption pairs \cre{largely} mined from the web \cre{(a small portion comes from the COCO dataset)}; thus, our model can be considered weakly supervised. Therefore, we \cre{also} compare against other weakly-supervised OCL methods. 
We further propose a variant termed \ourmethodDagger, which uses ground-truth class labels associated with the input image instead of the generated caption to generate and for semantic guidance. 

\subsection{Instance-aware object discovery}
The standard way to test how well the slots bind to an object is to evaluate on the object discovery task, \ie, producing a set of masks that cover the independent
objects appearing in an image.
We compare \ourmethod against existing SotA object-centric methods using the standard multi-object discovery metrics popular in the OCL literature \cite{wu2023slotdiffusion,seitzer2022bridging,jiang2023object}. This includes \emph{(i)} the mean Intersection over Union between the predicted masks from the slots, computed using the attention weights $\attn(\slot, \feat)$ as defined in \cref{eq:slot_attn}, and the ground-truth \emph{instance} masks, mIoU$_i$, 
\emph{(ii)} the mean Best Overlap over instance-level masks, $\text{mBO}_i$, and \emph{(iii)} over class-level masks, $\text{mBO}_c$.
Please see the supplement for details and additional results for the foreground-adjusted Rand index, FG-ARI. 
\Cref{tab:exp:od_cmp_ocl_only} shows that \ourmethod outperforms all previous OCL methods across mIoU$_{i}$, mBO$_i$, and mBO$_c$ metrics by a wide margin. 
\Cref{fig:qualitative:OD} shows qualitative results for object discovery compared to DINOSAUR~\cite{seitzer2022bridging}, StableLSD \cite{jiang2023object}, and SPOT \cite{kakogeorgiou2024spot}. 
They show that our method decomposes a scene in a more instance-centric way with sharper boundaries, no object splitting, and cleaner background segmentation. Importantly, unlike SPOT, our model can correctly segment different instances of the same object class, see \cref{fig:qualitative:OD}.
\input{tables/nips_tabs/sota_od_cmp_ocl}
\input{figures/od_qualitative}
\input{tables/nips_tabs/so_po_go}

\myparagraph{SO-PO-GO metrics.}
A major reason for our method's success is because it reduces the over- and under-segmentation (part-whole ambiguity) issues, which plague existing OCL methods. To quantify how well \ourmethod resolves these ambiguities, we evaluate using the SO-PO-GO metric proposed by \cite{fan2023unsupervised}. The metric reports the percentage of slots that bind to a single object (\emph{SO}), slots that bind to part of an object (\emph{PO}), and slots that bind to a group of objects (\emph{GO}). As seen in \cref{tab:so_po_go}, our method has a much higher percentage of slots that bind to a single object while reducing the number of slots that bind to parts of objects compared to SotA OCL methods. \cre{Our method also yields a higher percentage of background slots compared to other methods.}

\myparagraph{Zero-shot learning.}
We next show that reducing the part-whole ambiguity also helps object discovery (OD) in a zero-shot manner. As the slots are now biased towards objects, they can better segment scenes compared to baseline methods even when not trained on them. We take \ourmethod trained on the COCO dataset and report the zero-shot OD results on the CLEVRTex \cite{karazija2021clevrtex} and Obj365 \cite{shao2019objects365} datasets, see \cref{tab:abl4}. We obtained the masks for the Obj365 dataset by prompting SAMv2 \cite{ravi2024sam2} with ground-truth bounding boxes. 
We observe that our approach again outperforms SotA OCL methods.
\input{tables/nips_tabs/zero_shot}

\myparagraph{Comparison to weakly-supervised OCL.} 
\ourmethod{} can be considered weakly supervised due to its dependence on BLIP-2 \cite{li2023blip} for caption generation. We show that this form of (very) weak supervision performs much better than more expensive weak supervision signals, such as bounding boxes or knowing the number of objects in the scene. In particular, we compare our method against two weakly-supervised variants of StableLSD: 
\emph{(i)} StableLSD-BBox, which uses the bounding-box information associated with each object for initializing the slots. 
This form of guidance has been previously used in \cite{kipf2022conditional}. 
\emph{(ii)} StableLSD-Dynamic, which, instead of having a fixed number of slots for each scene, dynamically assigns each scene the number of slots equal to the number of objects present. 
\cite{zimmermann2023sensitivity} showed that this is useful for addressing the issue of part-whole ambiguity, leading to better object discovery. 
We choose StableLSD for comparison since it is closest to our model regarding the supported downstream tasks (see \cref{tab:rel_work:baselines}).
As seen in \cref{tab:exp:od_cmp_wsl_ocl}, the weakly-supervised variants of StableLSD outperform StableLSD. 
\emph{Importantly}, \ourmethod outperforms both weakly-supervised methods even though it uses an even weaker supervision signal. 
\input{tables/nips_tabs/cmp_wsl_ocl}

\input{tables/nips_tabs/ablations_arch}
\myparagraph{Importance of semantic and instance guidance.}
Next, we evaluate the contribution of the semantic and instance guidance losses.
\Cref{tab:abl0} shows the mIoU$_i$ metrics with different combinations of our three loss functions ($\mathcal{L}_\text{Recon},\ \mathcal{L}_{\text{Semantic}},\ \text{and}\ \mathcal{L}_{\text{Instance}}$).
We observe that combining semantic and instance losses together produces much better results than using them individually.
More importantly, the qualitative results in \cref{fig:ablation:loss} show that using only the reconstruction loss results in a noisy segmentation (over- and under-segmentation). Adding the semantic loss yields more precise boundaries, making the segmentation much less noisy. However, just using the semantic loss causes semantic drift and binds slots to semantic classes (under-segmentation); adding the instance guidance avoids the semantic drift problem and makes slots bind to objects instead of semantic classes. Thus, utilizing both semantic and instance guidance alleviates the over- and under-segmentation issue, yielding better slot embeddings for downstream tasks.

\myparagraph{Performance with different encoder networks.}
We next ablate the dependence of \ourmethod on the encoder architecture. We compare three different encoder models, namely Masked Auto Encoders (MAE) \cite{he2022masked}, DINOv2 \cite{oquab2023dinov2}, and DINOv1~\cite{caron2021emerging}. As seen in \cref{tab:abl1}, our method is robust to the choice of the encoder model. Moreover, it outperforms the model closest to ours in terms of downstream capabilities (StableLSD) for all encoder model architectures.
\input{figures/loss_ablation_cmp}

\myparagraph{Importance of pseudo ground-truth generation module.}
A key advantage of our method is utilizing the decoder model for both decoding the slots and also as semantic guidance generator, resulting in no additional dependency for guidance generation. 
We next show that our method of obtaining the guidance signal is superior to obtaining the guidance signal from models such as SAMv2 \cite{ravi2024sam2}.
To assess the impact of the semantic guidance signal, we turn off instance guidance ($\lambda_{i}\!=\!0)$ and set $\lambda_{s}\!=\!1$ for this experiment. 
As seen in \cref{tab:abl2}, our pseudo-ground truth signals lead to better results than using masks from SAMv2. This is because, without prompting, SAMv2 produces masks that are either over- or under-segmented compared to masks from our method.
To use SAMv2 effectively, we would need an additional prompt, \eg, a bounding box, but this form of supervision is more expensive than generated captions or image-level labels.

\subsection{Generative capabilities}
\paragraph{Conditional generation/reconstruction.}
\input{tables/nips_tabs/dpp_cmp}
Using a diffusion-based decoder in \ourmethod enables our model to conditionally generate the input image back from the slots and, more importantly, to be able to compositionally generate new scenes. 
We benchmark \ourmethod against StableLSD for conditional image generation, as this is the only OCL model to date to be able to reconstruct complex real-world images. We report the PSNR, SSIM \cite{wang2004image}, LPIPS \cite{zhang2018perceptual}, and FID \cite{heusel2017gans} metrics.
Both quantitatively (see \cref{tab:exp:fid}) and qualitatively (\cf \cref{fig:qual:CG}), our method outperforms StableLSD. The FID is calculated using 512 generated and real COCO images.
The qualitative results show that \ourmethod can reconstruct the input image more faithfully and with higher fidelity. 
\input{figures/fid_new_qual}

\myparagraph{Compositional generation.}
To the best of our knowledge, \ourmethod is the first slot-attention method to be able to compositionally generate complex real-world scenes with high fidelity. In \cref{fig:exp:comp_gen}, we show examples where objects can be removed from an input scene by removing a slot, or objects can be added to a scene by adding the slots from a different scene. Compositional generation with StableLSD results in very low-fidelity images, see supplement.
\input{figures/compositional_generation}
\input{tables/nips_tabs/dpp_acc}

\subsection{Property prediction}
\textbf{Instance-level property prediction} assesses the quality of the slot representation.
In this task, we predict object properties, such as class labels and object positions (centre of the object's bounding box) in the input images from the learned slot embeddings. We compare the informativeness of the features learned by slots of \ourmethod and StableLSD.
We report top-1 accuracy for label prediction and mean squared error for predicting the object's center.  
As seen in \cref{tab:exp:dpp}, \ourmethod consistently outperforms StableLSD for both tasks, indicating that our learned slots contain more informative features about the object than StableLSD's slot embeddings.

\subsection{Semantic-level object discovery}
Since our method makes use of large-scale pre-trained foundational models \cite{rombach2022high,oquab2023dinov2,caron2021emerging}, we also compare it against previous approaches for semantic-level segmentation \cite[\eg,][]{zhou2022extract,luo2023segclip,wu2023diffumask,couairon2024zeroshot}, utilizing features from these foundational models.
While our method is designed for instance-level segmentation, it can be modified to enable semantic-level segmentation. We consider a variant \cre{GLASS-Sem} (semantic-focused GLASS), where we purposefully under-segment the image (one slot is responsible for multiple objects belonging to the same class). 
For this, we set the instance guidance loss term to a low value ($\lambda_i\!=\!0.1$) during training. 
We report the mIoU$_c$ metric computed between the predicted masks from the slots and the ground-truth \emph{semantic} masks. 
\input{tables/nips_tabs/sota_od_cmp_all}

\cref{tab:exp:full_od_cmp} shows that our method outperforms not only all object-centric learning methods but also methods that rely on features from large-scale models for performing semantic-level object discovery.  We attribute the improvement of \ourmethod 
to its careful interplay of features between the different foundational models: 
Our approach aggregates features from a foundation model (DINOv2 \cite{oquab2023dinov2}) 
but this feature aggregation is guided by our semantic guidance module, which helps it achieve precise boundaries. 
This interpretation is supported by the observation that \ourmethod outperforms models such as Dataset Diffusion \cite{nguyen2023dataset} even though, just like \ourmethod, they use Stable Diffusion features for creating pseudo masks.

\myparagraph{Limitations.}
\cre{
Though our work achieves better results than SotA baseline OCL methods on several tasks, it still has several limitations: 
\emph{(i)} First, like most slot attention-based methods, our method is sensitive to the number of slots (\cf Tab.\ 12 in the supplement). 
\emph{(ii)} Though our method is more instance-focused, it sometimes binds slots to semantics and not instances (\cf Fig.\ 10 in the supplement). \emph{(iii)} The image generation results are still below SotA generative models and need further improvement.}

%% file: tables/nips_tabs/sota_od_cmp_ocl.tex
\begin{table}[tbp]
    \centering
    \scriptsize
    \setlength{\tabcolsep}{1pt}
    \smallskip    
    \begin{tabularx}{\columnwidth}
    {@{}
    >{\raggedright\arraybackslash}X
    >{\centering\arraybackslash}p{0.675cm}
    >{\centering\arraybackslash}p{0.675cm}
    >{\centering\arraybackslash}p{0.775cm}
    >{\centering\arraybackslash}p{0.675cm}
    >{\centering\arraybackslash}p{0.675cm}
    >{\centering\arraybackslash}p{0.775cm}
    @{}}
        \toprule
        Model 
        & \multicolumn{3}{c}{COCO (in \%, all $\uparrow$)}
        & \multicolumn{3}{c}{VOC (in \%, all $\uparrow$)}  \\
        \cmidrule(lr){2-4}
        \cmidrule(lr){5-7}
        & $\text{mIoU}_i$ 
        & $\text{mBO}_i$ 
        & $\text{mBO}_c$ 
        & $\text{mIoU}_i$ 
        & $\text{mBO}_i$ 
        & $\text{mBO}_c$ 
        \\
        \midrule
        SA$^*$ \cite{locatello2020object} \tiny{NeurIPS'20} & -- & 17.2 & 19.2 & -- & 24.6 & 24.9 \\
        SLATE$^*$ \cite{singh2021illiterate} \tiny{ICLR'22} & -- & 29.1 & 33.6 & -- & 35.9 & 41.5  \\
        DINOSAUR-MLP \cite{seitzer2022bridging} \tiny{ICLR'23} & 26.8 & 28.1 & 32.1 & 39.1 & 39.7 & 41.2 \\ 
        DINOSAUR-Trans. \cite{seitzer2022bridging} \tiny{ICLR'23} & 31.6 & 33.3 &  41.2 & 42.0 & 43.2 & 47.8 \\ 
        SPOT \cite{kakogeorgiou2024spot} \tiny{CVPR'24} & 34.0 & 35.0 & 44.7 & 48.8 & 48.3 & 55.6 \\ 
        SlotDiffusion$^*$ \cite{wu2023slotdiffusion} \tiny{NeurIPS'23} & -- & 31.0 & 35.0 & -- & 50.4 & 55.3 \\
        StableLSD \cite{jiang2023object} \tiny{NeurIPS'23} & 24.7 & 25.9 & 30.0 & 31.5 & 32.1 & 35.4 \\

        \rowcolor{gray!15} \ourmethodDagger (ours) & \best{39.0} \tiny{(+5.0)} & \best{40.8} \tiny{(+5.8)} & \best{48.7} \tiny{(+4.0)} & \secondbest{57.8} \tiny{(+9.0)} & \secondbest{58.5} \tiny{(+8.1)} & \secondbest{61.5} \tiny{(+5.9)} \\ 
        \rowcolor{gray!15} GLASS (ours) & \secondbest{38.9} \tiny{(+4.9)} & \secondbest{40.6} \tiny{(+5.6)} & \secondbest{48.5} \tiny{(+3.8)} & \best{58.1} \tiny{(+9.3)} & \best{58.9} \tiny{(+8.5)} & \best{62.2} \tiny{(+6.6)}\\ 
        \bottomrule
    \end{tabularx}
    \vspace{-0.75em} 
    \caption{\textbf{Comparison between OCL methods for instance-aware object discovery.} \ourmethod and \ourmethodDagger clearly outperform all other SotA OCL methods on the multi-object discovery metrics. The best value is highlighted in \best{bold}, the second best is \secondbest{underlined}. $^*$ numbers are taken from \cite{kakogeorgiou2024spot}. 
    Values in parentheses denote the improvement of \ourmethod over the previous SotA method. 
    See \cref{tab:exp:full_od_cmp} for additional information about each method, \eg, pre-trained models used, input modalities, and downstream capabilities.}
    \label{tab:exp:od_cmp_ocl_only}
    \vspace{-0.5em}
\end{table}

%% file: figures/od_qualitative.tex
\newcolumntype{Z}{>{\centering\arraybackslash}X}
\definecolor{mymethodcolor}{RGB}{101, 143, 255}
\newcommand{\mymethod}[1]{\textcolor{mymethodcolor}{\textbf{#1}}}

\begin{figure*}
    \scriptsize
    \setlength{\tabcolsep}{1pt}
    \centering
    \begin{tabularx}{\textwidth}{@{}
    *{6}{Z}|
    *{6}{Z}
    @{}}
    Input & DINOSAUR \cite{seitzer2022bridging} & StableLSD \cite{jiang2023object} & SPOT \cite{kakogeorgiou2024spot} & \textbf{GLASS (ours)} & \textbf{GLASS$^{\dagger}$ (ours)} & Input & DINOSAUR \cite{seitzer2022bridging} & StableLSD \cite{jiang2023object} & SPOT \cite{kakogeorgiou2024spot} & \textbf{GLASS (ours)} & \textbf{GLASS$^{\dagger}$ (ours)} \\
   \includegraphics[width=0.08\textwidth]{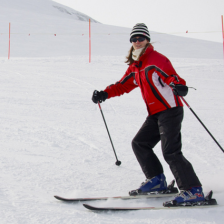} 
    &\includegraphics[width=0.08\textwidth]{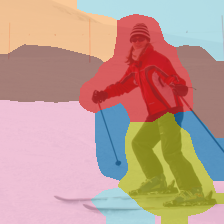} &\includegraphics[width=0.08\textwidth]{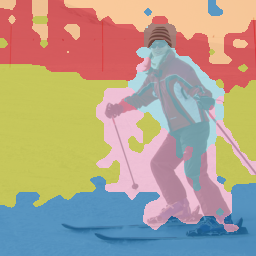}
    &\includegraphics[width=0.08\textwidth]{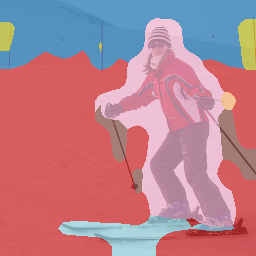}
    &\includegraphics[width=0.08\textwidth]{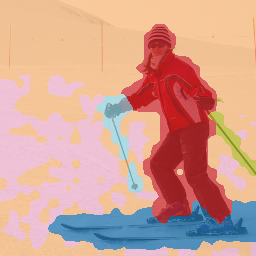}
    &\includegraphics[width=0.08\textwidth]{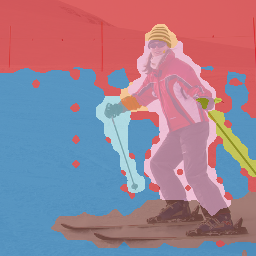} 

    &\includegraphics[width=0.08\textwidth]{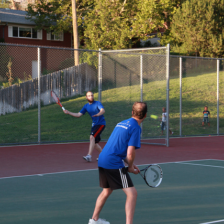} 
    &\includegraphics[width=0.08\textwidth]{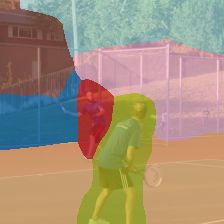} &\includegraphics[width=0.08\textwidth]{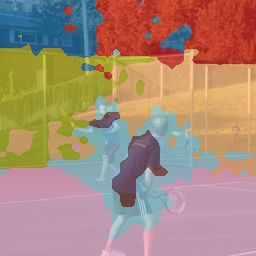}
    &\includegraphics[width=0.08\textwidth]{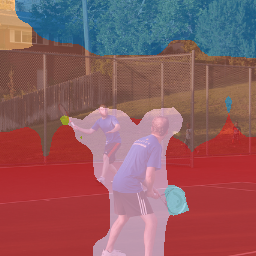}
    &\includegraphics[width=0.08\textwidth]{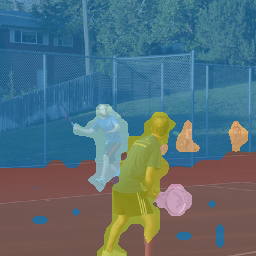}
    &\includegraphics[width=0.08\textwidth]{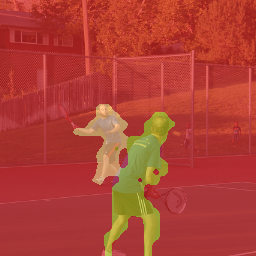}
     \\
    \includegraphics[width=0.08\textwidth]{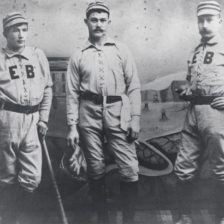} 
    &\includegraphics[width=0.08\textwidth]{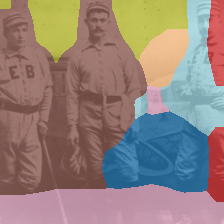} & \includegraphics[width=0.08\textwidth]{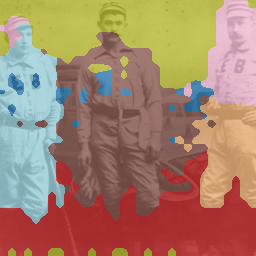} 
    & \includegraphics[width=0.08\textwidth]{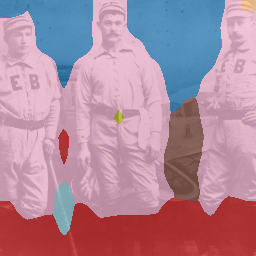} 
    & \includegraphics[width=0.08\textwidth]{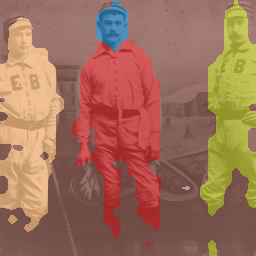} 
    & \includegraphics[width=0.08\textwidth]{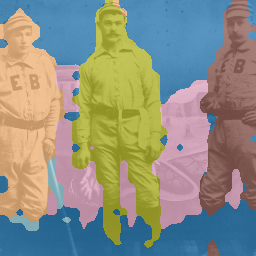} 
    & \includegraphics[width=0.08\textwidth]{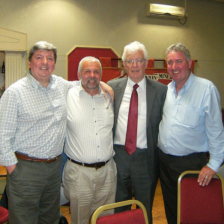} 
    &\includegraphics[width=0.08\textwidth]{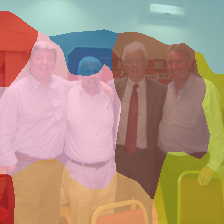} 
    &\scalebox{-1}[1]{
    \includegraphics[width=0.08\textwidth]{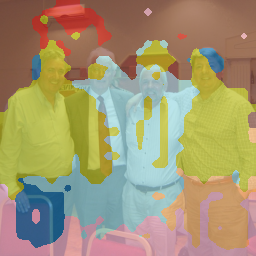}}
    &\includegraphics[width=0.08\textwidth]{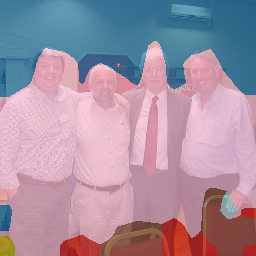}
    &\includegraphics[width=0.08\textwidth]{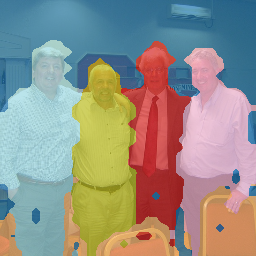}
    &\includegraphics[width=0.08\textwidth]{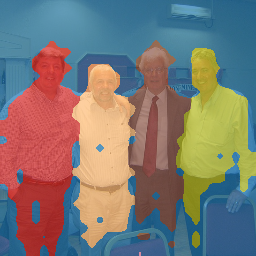}
    \\
    \end{tabularx}

    \vspace{-1em}
    \caption{\textbf{Qualitative comparison for object discovery.} \ourmethod and \ourmethodDagger can decompose an image at an instance level and reduces over- and under-segmentation of objects. Our method also yields cleaner boundaries for objects compared to StableLSD and DINOSAUR.} 
    \label{fig:qualitative:OD}
    \vspace{-0.75em}
\end{figure*}

%% file: tables/nips_tabs/so_po_go.tex
\begin{table}[tbp]
\centering
\scriptsize
\setlength{\tabcolsep}{3pt}
\begin{tabularx}{\columnwidth}{@{}Xcccccc@{}}
\toprule
Model 
& \multicolumn{3}{c}{COCO (in \%)}
& \multicolumn{3}{c@{}}{VOC (in \%)}\\
\cmidrule(lr){2-4} 
\cmidrule(lr){5-7} 
& SO $(\uparrow)$ & PO $(\downarrow)$ & GO $(\downarrow)$ & SO $(\uparrow)$ & PO $(\downarrow)$ & GO $(\downarrow)$ \\
\midrule
StableLSD \cite{jiang2023object} \tiny{NeurIPS'23} & 10.2 & 87.3 & 1.6 & 6.7 & 91.6 & 0.40 \\
DINOSAUR \cite{seitzer2022bridging} \tiny{ICLR'23} & 22.1 & 71.2 & 2.1 & 22.4 & 70.2 & 0.07 \\
SPOT \cite{kakogeorgiou2024spot} \tiny{CVPR'24} & 24.7 & 69.7 & \secondbest{0.01} & 26.2 & 65.0 & \best{0.00}\\
\rowcolor{gray!15} \ourmethodDagger  & \best{27.3} & \secondbest{49.6} & \secondbest{0.01} & \best{30.4} & \best{26.2} & 0.67 \\
\rowcolor{gray!15} \ourmethod & \secondbest{25.2} & \best{45.9} & \best{0.00} & \secondbest{26.7} & \secondbest{42.3} & \secondbest{0.01} \\
\bottomrule
\end{tabularx}
\vspace{-0.75em}

\caption{\textbf{SO-PO-GO metrics.} Our method has a higher \% of slots that bind to a single object compared to baselines, while also being less prone to over-segmentation and under-segmentation as seen by PO and GO metrics. \% of slots binding to background not shown.}
\label{tab:so_po_go}
\vspace{-0.5em}
\end{table}

%% file: tables/nips_tabs/zero_shot.tex
\begin{table}[tbp]
\scriptsize
\centering

\begin{tabularx}{\columnwidth}{@{}Xcccc@{}}
\toprule
Model 
& \multicolumn{2}{c}{CLEVRTex (in \%, all $\uparrow$)}
& \multicolumn{2}{c@{}}{Obj365 (in \%, all $\uparrow$)}\\
\cmidrule(lr){2-3} 
\cmidrule(lr){4-5} 
& mIoU$_i$ & mBO$_i$ & mIoU$_i$ & mBO$_i$ \\
\midrule

StableLSD \cite{jiang2023object} \tiny{NeurIPS'23} & 24.0 & 27.6 & 14.8 & 16.9 \\
DINOSAUR \cite{seitzer2022bridging} \tiny{ICLR'23} & 30.2 & 35.1 & 16.2 & 18.9 \\
SPOT \cite{kakogeorgiou2024spot} \tiny{CVPR'24} & 39.5 & 43.7 & 18.0 & 20.7 \\
\rowcolor{gray!15}\ourmethodDagger & \best{47.2} & \best{52.3} & \best{19.6} & \best{22.4} \\
\rowcolor{gray!15}\ourmethod & \secondbest{46.1} & \secondbest{50.1} &  \secondbest{18.6} & \secondbest{21.4}\\

\bottomrule
\end{tabularx}
\vspace{-.75em}

\caption{
\textbf{Zero-shot object discovery}. Our method outperforms the baseline methods on the task of zero-shot object discovery.}
\label{tab:abl4}
\vspace{-0.5em}
\end{table}

%% file: tables/nips_tabs/cmp_wsl_ocl.tex
\begin{table}[tbp]
    \centering
    \scriptsize
    \setlength{\tabcolsep}{4pt}
    \begin{tabularx}{\columnwidth}
    {@{}
    >{\raggedright\arraybackslash}X
    >{\centering\arraybackslash}p{0.75cm}
    >{\centering\arraybackslash}p{0.75cm}
    >{\centering\arraybackslash}p{0.75cm}
    @{}}
        \toprule
        Model 
        & \multicolumn{3}{c}{VOC (in \%, all $\uparrow$)} \\
        \cmidrule(lr){2-4}
        & $\text{mIoU}_i$ 
        & $\text{mBO}_i$ 
        & $\text{mBO}_c$ \\
        \midrule
        
        StableLSD \cite{jiang2023object} \tiny{NeurIPS'23} 
        & 31.5 & 32.1 & 35.4 \\
        StableLSD-Bbox & 31.5 & 38.4 & 43.0 \\ 
        StableLSD-Dynamic & 30.0 & 37.3 & 42.3 \\ 
        \rowcolor{gray!15} \ourmethodDagger (ours) &  \secondbest{57.8} & \best{58.5} & \secondbest{61.5}\\ 

        \rowcolor{gray!15} GLASS (ours) & \best{58.1} & \secondbest{58.9} & \best{62.2} \\ 
        \bottomrule
    \end{tabularx}
    \vspace{-0.75em} 
    \caption{\textbf{Comparison with weakly-supervised baselines,} \ie, variants of StableLSD. \ourmethod clearly outperforms the weakly-supervised variants of the StableLSD model even though it uses weaker supervision than these variants.}
    \label{tab:exp:od_cmp_wsl_ocl}
    \vspace{-1em}
\end{table}

%% file: tables/nips_tabs/ablations_arch.tex
\begin{table*}[htbp]
\scriptsize
\subfloat[
\textbf{Importance of semantic and instance guidance losses}. 
A combination of semantic and instance loss terms performs the best.
\label{tab:abl0}
]{
\begin{minipage}[t]{0.31 \linewidth}{
\setlength{\tabcolsep}{2pt}
\begin{tabularx}{\textwidth}{@{}X@{}cc@{}}
\toprule
Loss term
& \multicolumn{2}{@{}c@{}}{mIoU$_i$ (in \%, $\uparrow$)}\\
\cmidrule{2-3} 
& COCO & VOC \\
\midrule
$\mathcal{L}_{\text{Recon}}$ & 30.0 &  34.4 \\
$\mathcal{L}_{\text{Recon}} + 0.7 \mathcal{L}_{\text{Semantic}}$ & 30.9 & 55.1 \\
$\mathcal{L}_{\text{Recon}} + 0.9 \mathcal{L}_{\text{Instance}}$ & 29.3 & 38.9 \\
$\mathcal{L}_{\text{Recon}} + 0.7 \mathcal{L}_{\text{Semantic}} +  0.9 \mathcal{L}_{\text{Instance}}$ & \textbf{38.9} & \textbf{58.1} \\
\bottomrule
\end{tabularx}
}\end{minipage}
}
\hfill
\subfloat[
\textbf{Effect of encoder network.} \ourmethod is robust to the encoder architecture and outperforms the baseline even with weaker encoder networks.
\label{tab:abl1}
]{
\begin{minipage}[t]{0.31\linewidth}{
\begin{tabularx}{\textwidth}{@{}X@{}cc@{}}
\toprule
Encoder 
& \multicolumn{2}{@{}c@{}}{mIoU$_i$ (in \%, $\uparrow$)}\\
\cmidrule{2-3} 
& COCO & VOC  \\
\midrule
Baseline (StableLSD \cite{jiang2023object}) & 24.7 & 31.5 \\
\ourmethod w/ MAE \cite{he2022masked} & 30.0 & 41.1 \\
\ourmethod w/ DINOv1 \cite{caron2020unsupervised} & 31.4 & 54.2 \\
\ourmethod w/ DINOv2 \cite{oquab2023dinov2} & \textbf{38.9} & \textbf{58.1} \\
\bottomrule
\end{tabularx}
}\end{minipage}
}
\hfill%
\subfloat[
\textbf{Effectiveness of pseudo GT semantic mask}.
Using masks from the decoder performs better than masks obtained from SAMv2.
\label{tab:abl2}
]{
\begin{minipage}[t]{0.31\linewidth}{
\begin{tabularx}{\textwidth}{@{}X@{}cc@{}}
\toprule
Model
& \multicolumn{2}{@{}c@{}}{mIoU$_i$ (in \%, $\uparrow$)}\\
\cmidrule{2-3} 
& COCO & VOC  \\
\midrule
\ourmethod w/ SAMv2 & 30.4 & 42.1 \\
\ourmethodDagger ($\lambda_s\!=\!1, \lambda_i\!=\!0$) & 31.1 & \textbf{58.6} \\
\ourmethod ($\lambda_s\!=\!1, \lambda_i\!=\!0$) & \textbf{32.2} & 55.6 \\
\bottomrule
\end{tabularx}
\vspace{6pt}
}\end{minipage}
}

\vspace{-.5em}
\caption{
\textbf{Ablation study.}
\emph{(a)} We study the impact of different loss terms on \ourmethod, \emph{(b)} the impact of different encoder architectures, and \emph{(c)} the impact of using different guidance generation on the performance of our approach on the instance-level object discovery task.
}
\label{tab:ablations} 
\vspace{-1em}
\end{table*}

%% file: figures/loss_ablation_cmp.tex
\begin{figure}
    \scriptsize 
    \setlength{\tabcolsep}{1pt}
    \centering
    \begin{tabular}{@{}*{4}{x{54}}@{}}
    $\mathcal{L_{\text{Recon}}}$ & $\mathcal{L}_{\text{Recon}} + \mathcal{L}_{\text{Semantic}}$ & $\mathcal{L}_{\text{Recon}} + \mathcal{L}_{\text{Instance}}$ & \textbf{\ourmethod (ours)} \\[1pt]
    \includegraphics[width=0.10\textwidth]{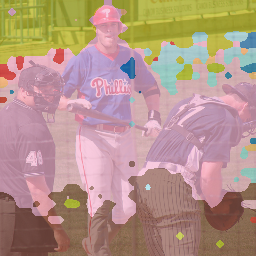} &
    \includegraphics[width=0.10\textwidth]{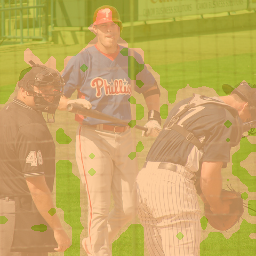} & 
    \includegraphics[width=0.10\textwidth]{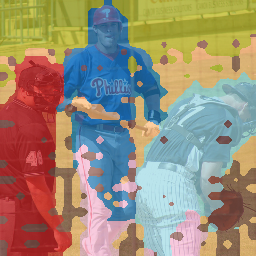} & 
    \includegraphics[width=0.10\textwidth]{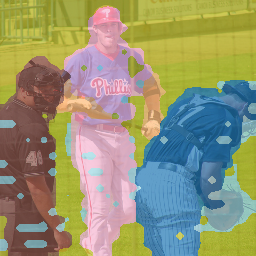} \\
    \includegraphics[width=0.10\textwidth]{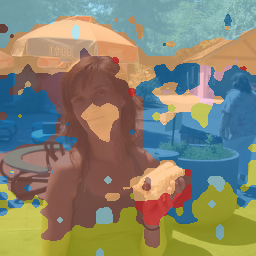} &
    \includegraphics[width=0.10\textwidth]{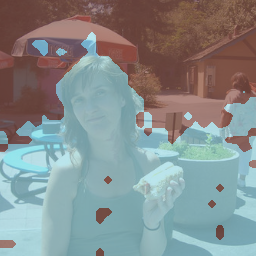} & 
    \includegraphics[width=0.10\textwidth]{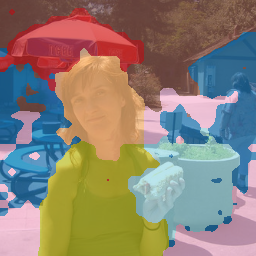} & 
    \includegraphics[width=0.10\textwidth]{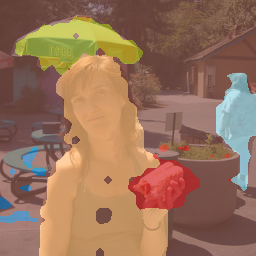} \\
    \end{tabular}
    \vspace{-0.75em}
    \caption{\textbf{Qualitative results showing the importance of joint semantic and instance guidance.} Using both guidances together provides precise boundaries and biases the slots to object instances.} 
    \label{fig:ablation:loss}
    \vspace{-0.5em}
\end{figure}

%% file: tables/nips_tabs/dpp_cmp.tex
\begin{table}
    \centering
    \scriptsize
    \setlength{\tabcolsep}{4pt}
    \smallskip
    \begin{tabularx}{\columnwidth}{@{}Xcccc@{}}
        \toprule
         Model 
        & PSNR (in dB, $\uparrow$) & SSIM ($\uparrow$) & LPIPS ($\downarrow$) & FID ($\downarrow$) \\
        \midrule
        StableLSD \cite{jiang2023object} \tiny{NeurIPS'23} & 10.92 & 0.20 & 0.72 & 140.62 \\
        \rowcolor{gray!15 }\ourmethodDagger (ours) & 10.88 & 0.20 & \textbf{0.59} & 79.61  \\
        \rowcolor{gray!15} \ourmethod (ours) & \textbf{10.93} & \textbf{0.21} & \textbf{0.59} & \textbf{71.30} \\
        \bottomrule
    \end{tabularx}
    \vspace{-0.75em}

    \caption{\textbf{Conditional generation.} Comparison between StableLSD and our approach for the conditional generation / recon.\ task.}
    \label{tab:exp:fid}   
    \vspace{-0.5em}
\end{table}

%% file: figures/fid_new_qual.tex
\begin{figure}
    \centering
    \scriptsize
    \setlength{\tabcolsep}{1pt}
    \begin{tabular}{@{}*{4}{x{54}}@{}}
    Original & StableLSD \cite{jiang2023object} & \textbf{\ourmethod$^{\dagger} $ (ours)} & \textbf{\ourmethod (ours)} \\
    \includegraphics[width=0.10\textwidth]{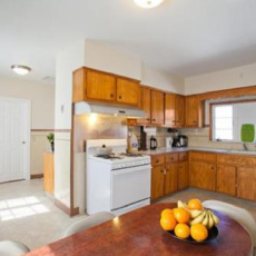} &
    \includegraphics[width=0.10\textwidth]{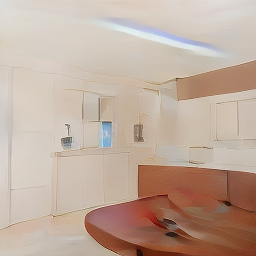} &
    \includegraphics[width=0.10\textwidth]{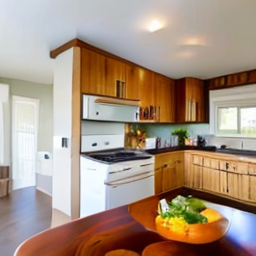} &
    \includegraphics[width=0.10\textwidth]{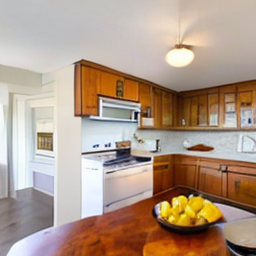} \\
    \includegraphics[width=0.10\textwidth]{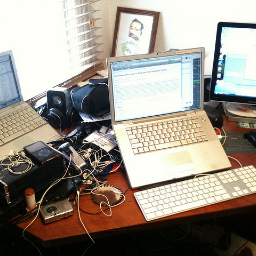} &
    \includegraphics[width=0.10\textwidth]{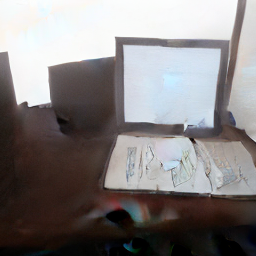} &
    \includegraphics[width=0.10\textwidth]{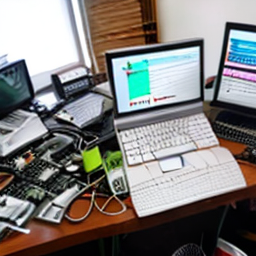} &
    \includegraphics[width=0.10\textwidth]{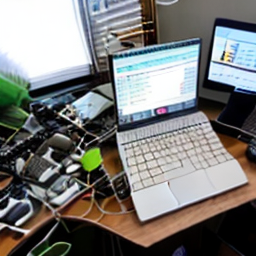} \\
    \end{tabular}
    \vspace{-0.75em}
    \caption{\textbf{Qualitative comparison for conditional image generation.} \ourmethod and \ourmethodDagger reconstruct the input scene more faithfully with a high degree of detail as compared to StableLSD.}
    \label{fig:qual:CG}
    \vspace{-0.5em}
\end{figure}

%% file: figures/compositional_generation.tex
\begin{figure}
    \centering
    \footnotesize
    \setlength{\tabcolsep}{1pt}
    \begin{tabular}{@{}
    *{3}{x{60}} 
    @{}}
    Remove item  & Original image & Edited image  \\
    \includegraphics[width=0.10\textwidth]{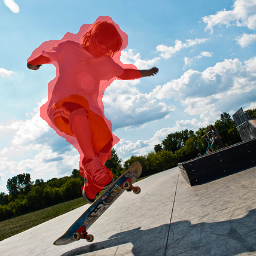}
    & \includegraphics[width=0.10\textwidth]{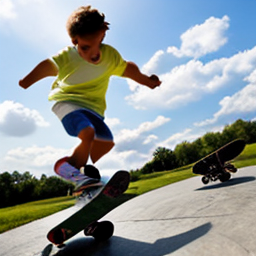}
    & \includegraphics[width=0.10\textwidth]{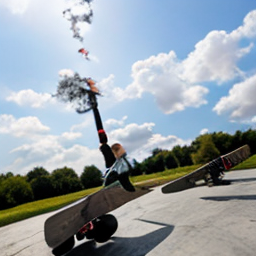} \\
    Add item  & Original image & Edited image \\
    \includegraphics[width=0.10\textwidth]{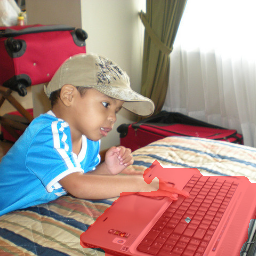}
    & \includegraphics[width=0.10\textwidth]{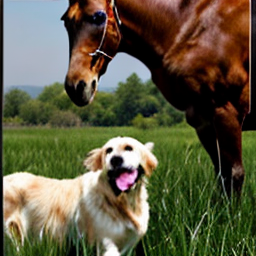}
    & \includegraphics[width=0.10\textwidth]{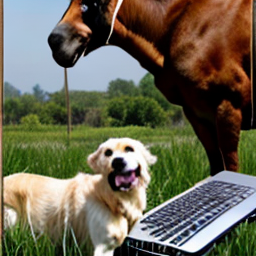} \\
    \end{tabular}
    \vspace{-.75em}
    \caption{\textbf{Compositional generation.} \ourmethod enables compositional image generation of real-world complex scenes. Here, the masked object \emph{(in red)} is the slot to be removed from or added to the original image. Please see supplement for more results.}
    \label{fig:exp:comp_gen}
    \vspace{-0.5em}
\end{figure}

%% file: tables/nips_tabs/dpp_acc.tex
\begin{table}
    \centering
    \scriptsize
    \setlength{\tabcolsep}{3pt}
    \smallskip

    \begin{tabularx}{\columnwidth}{@{}Xcccc@{}}
        \toprule
        Model 
        & \multicolumn{2}{c}{VOC}
        & \multicolumn{2}{c@{}}{COCO}\\
        \cmidrule(lr){2-3}
        \cmidrule(lr){4-5}
        & Acc (in \%, $\uparrow$) & MSE ($\downarrow$) & Acc (in \%, $\uparrow$) & MSE ($\downarrow$) \\
        \midrule
        StableLSD \cite{jiang2023object} \tiny{NeurIPS'23} & 55.1 & 0.039 & 16.4 & 0.062 \\
        \rowcolor{gray!15} \ourmethod (ours) & \textbf{58.1} & \textbf{0.037} & \textbf{20.8} & \textbf{0.059} \\
        \bottomrule
    \end{tabularx}
    \vspace{-0.5em}
    \caption{\textbf{Instance-level property prediction.} Comparison between StableLSD and \ourmethod for the property prediction task.}
    \label{tab:exp:dpp}
    \vspace{-0.5em}
\end{table}

%% file: tables/nips_tabs/sota_od_cmp_all.tex
\newcommand{\img}{$\mathcal{I}$}
\newcommand{\capt}{$\mathcal{C}$}
\newcommand{\mylabel}{$\mathcal{L}$}

\begin{table}[t]
    \smallskip

    \centering
    \scriptsize
    \setlength{\tabcolsep}{2pt}
    \renewcommand{\arraystretch}{1.10}
    \begin{tabularx}{\columnwidth}
    {@{}
    >{\raggedright\arraybackslash}X
    >{\raggedright\arraybackslash}p{1.25cm}
    >{\raggedright\arraybackslash}p{0.75cm}
    >{\raggedright\arraybackslash}p{1.5cm}
    >{\centering\arraybackslash}p{0.75cm}
    >{\centering\arraybackslash}p{0.75cm}
    @{}}
    \toprule
    Model & \multirow{2}{=}{Downstr.\
    tasks} & Input & \multirow{2}{=}{Pre-trained models} &  \multicolumn{2}{c}{$\text{mIoU}_c$ (in \%, $\uparrow$)} \\
    \cmidrule{5-6}
    &
    &
    &
    & COCO 
    & VOC
    \\
    \midrule
  1Stage \cite{araslanov2020single} \tiny{CVPR’20} & sOD & \img{} + \mylabel{} & -- & -- & 62.7 \\ 
  AFA \cite{ru2022learning} \tiny{CVPR’22} & sOD & \img{} + \mylabel{} & -- & 38.9 & 66.0 \\
  ToCo \cite{ru2023token} \tiny{CVPR’23} & sOD & \img{} + \mylabel{} & -- & 41.3 & \best{69.8} \\
    \midrule
    MaskCLIP \cite{zhou2022extract} \tiny{ECCV'22}                          & sOD & \img{} + \capt  & \tiny{CLIP} & 20.6 & 38.8 \\
    SegCLIP \cite{luo2023segclip} \tiny{ICML'23}                            & sOD & \img{} + \capt & \tiny{CLIP} & 26.5 & 52.6  \\
    CLIPPy \cite{ranasinghe2023perceptual} \tiny{ICCV'23}                   & sOD & \img{} + \capt & \tiny{CLIP} & 32.0 & 52.2 \\ 
    OVSeg \cite{xu2023learning} \tiny{CVPR'23}                              & sOD & \img{} + \capt  & \tiny{CLIP}  & 25.1 & 53.8 \\ 
    $\text{DeepSpectral}$ \cite{melas2022deep} \tiny{CVPR'22} & sOD & \img  &  \tiny{DINO} & -- & 37.2 \\ 
    $\text{COMUS}$ \cite{zadaianchuk2023unsupervised} \tiny{ICLR'23}               & sOD & \img & \tiny{DINO} & -- & 50.0\\ 
    DiffuMask \cite{wu2023diffumask} \tiny{NeurIPS'23}                      & sOD & \img{} + \capt & \tiny{SD + CLIP + \cite{ahn2018learning}} & -- & 57.4 \\
    Dataset Diffusion \cite{nguyen2023dataset} \tiny{NeurIPS'23}            & sOD & \img{} & \tiny{SD} + \tiny{BLIP-2} & 34.2 & 64.8
    \\ 
   \midrule
   DiffCut \cite{couairon2024zeroshot} \tiny{NeurIPS'24} & sOD & \img{} & \tiny{SD} & 34.1 & 65.2 \\
    OVDiff \cite{karazija2024diffusion} \tiny{ECCV'24}                    & sOD & \img{} + \capt & \tiny{SD + DINO + CLIP} & 34.6 & 66.3  \\
   EmerDiff \cite{namekata2024emerdiff} \tiny{ICLR'24}                  & sOD & \img & \tiny{SD} & 33.1 & 40.3  \\
    \midrule
    DINOSAUR-MLP \cite{seitzer2022bridging} \tiny{ICLR'23}                  & iOD + PP & \img & \tiny{DINO} & 31.7  & 41.0\\ 
    DINOSAUR-Transformer \cite{seitzer2022bridging} \tiny{ICLR'23}          & iOD + PP & \img  & \tiny{DINO} & 40.6 & 47.5 \\ 
      SPOT \cite{kakogeorgiou2024spot} \tiny{CVPR'24}          & iOD + PP & \img  & \tiny{DINO} & \secondbest{44.6}  & 55.3 \\ 
    StableLSD \cite{jiang2023object} \tiny{NeurIPS'23}                    & OD + PP + CG & \img & \tiny{SD + DINOv2} & 29.5 & 35.4\\ 
    \rowcolor{gray!15} GLASS-Sem (ours)                                       & iOD + PP + CG + CPG & \img{} & \tiny{SD + DINOv2 + BLIP-2} & \best{46.7} \tiny{(+2.1)} & \secondbest{68.9} \tiny{(-0.9)}  \\
    \bottomrule
    \end{tabularx}
    \vspace{-0.75em}
    \caption{
    \textbf{Comparison on semantic-level object discovery.}
    We compare GLASS-Sem ($\lambda_i\!=\!0.1, \lambda_s\!=\!1)$ with baselines for semantic-level object discovery. We divide the baselines into 4 different sections \emph{(i)} weakly-supervised, \emph{(ii)} training-based, \emph{(iii)} training-free, and \emph{(iv)} OCL methods. 
    \emph{Downstream tasks} denote a model's capability of solving the following tasks: 
    iOD /sOD -- instance-/semantic-level object discovery, PP -- instance-level property prediction, CG -- conditional generation, and CPG -- compositional generation.
    \emph{Input} denotes the input signal the model itself trains on, where \img{} -- image, \capt{} -- captions, and \mylabel{} -- image-level labels.
    \emph{Pre-trained models} denote the underlying foundation models used in the method. 
    \textbf{Note:} \emph{\ourmethod is not designed for sOD, but it is controllable and can be tuned for either the sOD or iOD task.}
    }
    \label{tab:exp:full_od_cmp}
    \vspace{-0.75em}
\end{table}

%% file: sec/conclusion.tex
\section{Conclusion}
We present \ourmethod, a novel object-centric learning method that learns in the space of generated images from a pre-trained diffusion model. 
Our method makes use of semantic and instance guidance in order to learn better instance-centric representations. We clearly outperform previous SotA OCL methods on various tasks: instance-level (zero-shot) object discovery and conditional image generation. Our work also surpasses SotA models that use large-scale pre-trained models for semantic-level object discovery, and learns better slot representations for instance-level property prediction than similarly versatile OCL methods.
Notably, our method is the first OCL approach that enables the compositional generation of complex real-world scenes.

{\small
\paragraph{Acknowledgements.}
\cre{
This project has received funding from
the European Research Council (ERC) under the European
Union’s Horizon 2020 program (grant agreement No.\ 866008).
The project was also supported in part by the State of Hesse
through the project “The Third Wave of Artificial Intelligence
(3AI)”.}}

%% file: suppl.tex
\def\authorstep{\hspace{0.75cm}}
\def\affiliationstep{\hspace{0.5cm}}
\clearpage
\setcounter{section}{0}
\twocolumn[{
\renewcommand\twocolumn[1][]{#1}
\maketitlesupplementary
{\large Krishnakant Singh\textsuperscript{1}
\authorstep Simone Schaub-Meyer\textsuperscript{1,2}
 \authorstep Stefan Roth\textsuperscript{1,2}\\[-4pt]
\textsuperscript{1}Department of Computer Science, TU Darmstadt\affiliationstep \textsuperscript{2}hessian.AI}
\vspace{0.75cm}
}]
In this appendix, we give additional details for purposes of reproducible research and show further results that provide more insights into our proposed \ourmethod{} method.
Our code is available at \href{https://github.com/visinf/glass/}{https://github.com/visinf/glass/}.

\section{Datasets} 
As in previous work \cite{jiang2023object,seitzer2022bridging,wu2023slotdiffusion}, we report all our results on the VOC \cite{everingham2010pascal} and COCO \cite{lin2014microsoft} datasets.
Both these datasets serve as popular benchmarks for object discovery and have been used to evaluate various object-centric learning methods on real-world images \cite{seitzer2022bridging,wu2023slotdiffusion,jiang2023object}. 

\myparagraph{VOC.} The PASCAL VOC dataset \cite{everingham2010pascal} is a standard dataset used in object discovery. We use the ``trainaug'' variant for generating the images and their corresponding mask. ``trainaug'' is an unofficial split of datasets, consisting of 10,582 images, which include 1,464 images from the original VOC segmentation train set and 9,118 images from the SBD dataset \cite{hariharan2011semantic}. 
\ourmethod and \ourmethod{}$^\dagger$ are evaluated on the official VOC validation set of 1,449 images 

\myparagraph{COCO.} The COCO dataset \cite{lin2014microsoft} consists of 118,287 images of complex multi-object scenes.
Unlike the VOC dataset, where images often contain only a single object in the scene, COCO images contain at least two objects, often even a dozen.
\ourmethod and \ourmethod{}$^\dagger$ are evaluated on a validation set with 5,000 COCO images.  

\input{tables/nips_tabs/training_details}
\section{Training Details} \label{sec:appendix:training_details}
\paragraph{Training dataset.} 
GLASS and GLASS$^\dagger$ are trained on images generated with a Stable Diffusion v2.1 \cite{rombach2022high} model. 
For training the model on the COCO-generated dataset, we generate 100K images and their corresponding pseudo ground-truth, obtained using COCO images and following the process in Sec.\ \ref{fig:method:network} of the main paper.
For training the model on the VOC-generated dataset, we generate approximately 10K images and their corresponding pseudo masks, obtained using VOC images following the process in Sec.\ \ref{fig:method:network} of the main paper.

\myparagraph{Architecture details.} 
\ourmethod for the COCO and VOC datasets is trained on a single NVIDIA A6000 Ada GPU. 
The training time for COCO models is typically 4 days, while for the VOC dataset, training is completed within two days. 
\ourmethod and its variants use DINOv2~\cite{oquab2023dinov2} with ViT-B \cite{kipf2022conditional} and a patch size of 14 as its encoder model, and Stable Diffusion (SD) v2.1 \cite{rombach2022high} as well as a three-layer MLP network as its decoder models. 

We train our model on the generated images in two phases.
In phase 1, only the slot attention module and the MLP decoder are trained with an Adam \cite{kingma2014adam} optimizer with a constant learning rate of 2e-5. In phase 2, the slot attention and MLP decoder module are trained with a learning rate of 1e-8 (essentially frozen). At the same time, we train the diffusion decoder with a learning rate of 4e-5 for the last 100K iterations.

\Cref{tab:appendix:training_detail} shows additional common details about the hyper-parameters and modules used in \ourmethod and \ourmethodDagger. 
When training on COCO-generated images, we train the model for 500K iterations, while when training on VOC-generated images, we train for 250K iterations. 
For the slot-attention module, the number of slot iterations in the GRU module is set to 5, and the number of slots is set to 7 for the COCO-generated and the VOC-generated dataset for \ourmethod. The slot size is set to 768; this configuration is akin to StableLSD \cite{jiang2023object}. The number of heads in the slot-attention module is set to 1, and a hidden size of 768 is used for the MLP. The final MLP layer in the slot-attention module projects the slots to a dimension of 1024.

\myparagraph{Pseudo labels.}
While generating each training image, we used Stable Diffusion's cross-attention and self-attention modules to extract the pseudo masks for the respective image (as described in Sec.\ \ref{sec:gsa} of the main paper; we used the same setup as in \cite{nguyen2023dataset}). We used a range-based thresholding to binarize these masks.
Specifically, we assign a pixel to the background if its objectness score (the max value among all class scores of $\mathbf{M}_{\text{ref}}$, see Sec.\ \ref{sec:gsa}) is below 0.4; conversely, if this objectness score is above 0.6, we assign the pixel to the foreground with the class label that has the max value. If the objectness score lies between 0.4 and 0.6, we assign a pseudo label of 255 (indicating that we are uncertain about the class label). This uncertain region helps in only calculating the semantic loss on regions with high certainty, avoiding uncertain regions.

\subsection{Object-level property  prediction}\label{sec:appendix:dpp_details}
For the object-level property prediction task, we train a single-layer linear model for label prediction and a single-layer linear network for position prediction. 
We use early stopping with a patience level set to 5 and train for 50 epochs on the VOC dataset and 10 epochs on the COCO training sets.
For matching the labels to the correct slots during training, we utilize the idea from 
\cite{dittadi2021generalization} and use the mIoU criterion for matching labels to slots.
Since the VOC training dataset does not have instance masks, we compute the IoU criterion using the semantic mask during training \cite{seitzer2022bridging}.
We use an AdamW \cite{loshchilov2017fixing} optimizer with a constant learning rate of 3e-4.
For the position prediction task, we normalize the image coordinates to lie between 0 and 1 by dividing the image coordinates by the image size. 

\input{tables/nips_tabs/comparision_fgari}
\section{Comparison with FG-ARI Scores}
Previous work in object-centric learning has regularly considered the FG-ARI metric for evaluation. The FG-ARI metric is a version of the adjusted Rand index (ARI) \cite{hubert1985comparing,rand1971objective}, which measures the similarity between two different clusterings in a permutation-invariant way by taking into account the foreground regions in the image.
However, the FG-ARI score is known to be an unreliable metric as discussed in \cite{kakogeorgiou2024spot,jiang2023object,karazija2021clevrtex,engelcke2020genesis}; it does not take into account background pixels and does not account for the shape of predicted masks. Please see \cite{kakogeorgiou2024spot} for further discussion. 
For completeness, we nevertheless provide results for the FG-ARI metric in \cref{tab:supp:fgari}.
While our FG-ARI scores are lower than some baselines, particularly on the COCO dataset, we believe that this should be mostly discounted due to the known deficiencies of this evaluation metric.
That said, our method is \emph{only} behind DINOSAUR on the VOC dataset and performs close to SPOT on the COCO dataset for the FG-ARI metric.
Additionally, we refer to the comprehensive results for the mIoU and mBO metrics in Tab.\ \ref{tab:exp:od_cmp_ocl_only} and the SO-PO-GO metrics in Tab.\ \ref{tab:so_po_go} of the main paper.

\input{figures/sam2_appx}
\section{Guidance with SAMv2 Masks}
Automatic segmentation using the SAMv2 model \cite{ravi2024sam2} requires careful tuning of many hyperparameters. Using default parameters results in severe under-segmentation issues (\cf \cref{fig:appendix:sam_failure}, where the humans are segmented as backgrounds). If we use a denser sampling of points, this results in an over-segmentation of objects into their parts, see \cref{fig:appendix:sam_failure}. These issues make plain SAMv2 segmentation masks unsuitable as guiding signals. We could overcome these issues with additional input prompts, such as bounding boxes, but this would make the annotation cost higher than simply using generated captions or image-level labels. 

\input{figures/compo_gen_stablelsd}
\section{Compositional Generation with StableLSD}

StableLSD \cite{jiang2023object} struggles with compositional generation, as shown in \cref{fig:appendix:comp_gen_lsd}. StableLSD is not able to add or remove objects from the original image; moreover, the quality of reconstruction and faithfulness of the input image is rather poor for StableLSD compared to results of our method, as shown in  \cref{fig:appendix:comp_gen_remove,fig:appendix:comp_gen_add}.

\section{Additional Results}
\paragraph{Additional qualitative results for object discovery.} \label{sec:appendix:qualitative_addn}
\cref{fig:appendix:OD} shows additional qualitative results for object discovery. \ourmethod decomposes the scene more cleanly and meaningfully than SotA OCL methods such as DINOSAUR \cite{seitzer2022bridging}, StableLSD \cite{jiang2023object}, and SPOT \cite{kakogeorgiou2024spot}. 
Images segmented by \ourmethod{} have more precise boundaries, the background segmentation is much cleaner, and the segmented regions do not split or group objects. 

\textbf{Additional comparison to DatasetDiffusion.}
Going beyond Tab.\ \ref{tab:exp:full_od_cmp} in the main paper, we additionally compare \ourmethod{} to an extended variant of DatasetDiffusion \cite{nguyen2023dataset}, which is trained with our 100K generated images for the COCO dataset and uses a DINOv2 backbone network. This extended variant of \cite{nguyen2023dataset} achieves an mIoU$_c$ of 42.8\,\%, while GLASS-Sem still achieves a higher score of 46.7\,\% on the COCO dataset.
This suggests that our proposed architecture and training scheme contribute significantly to the gains over DatasetDiffusion \cite{nguyen2023dataset}.

\myparagraph{Additional conditional generation results.}\label{sec:appendix:cond_gen}
\cref{fig:appendix:CG} shows additional results for the conditional generation of images using StableLSD, \ourmethodDagger, and \ourmethod. 
Our method generates images that are more faithful to the input image and have higher fidelity than StableLSD. 

\input{figures/compo_gen_cmp_comp_diffusion}
\input{figures/od_qualitative_appx}

\myparagraph{Additional compositional generation results.}
\cref{fig:appendix:comp_gen_remove,fig:appendix:comp_gen_add} show results from \ourmethod{} for compositional generation. In particular, we can see that we can add objects from one scene to another. This is possible even when the context is quite different, for example, adding a baseball player to the bowl of food (\cref{fig:appendix:comp_gen_add} \emph{(row-1)}). 
Also, we can remove objects completely from a scene. Please note that, to our knowledge, no other OCL method can perform these actions with this fidelity or faithfulness.

\afterpage{\clearpage}
\input{figures/fid_qual_appx}
\input{figures/compo_gen_remove_appx}
\input{figures/compo_gen_add_appx}

\myparagraph{Comparison to text-based compositional models.}
We conduct a preliminary study of comparing \ourmethod for compositional generation against a text-based compositional generation method, namely Composable Diffusion \cite{liu2022compositional}, which composes objects/concepts via text-based prompts. For example, ``class label 1 | class label 2'' generates an image containing class labels 1 and 2.
On the other hand, GLASS is a compositional method that first extracts objects/concepts from an image and then generates the image. 
\textbf{Note}: Both these models try to address the compositional generation problem. However, they are not directly comparable as \ourmethod relies on input images for extracting concepts while Composable Diffusion uses the text.
\cref{fig:appendix:comp_gen_cmp_comp_diffusion} shows that \ourmethod can extract concepts/slots from one image and transfer them to another to create a high-fidelity image, which faithfully contains both concepts from the source and destination image. In contrast, \cite{liu2022compositional} sometimes is unable to compose certain concepts given in the text prompt, \eg, in \cref{fig:appendix:comp_gen_cmp_comp_diffusion} (\emph{top-row}) no person is generated for the prompt ``PERSON AND PIZZA".

\section{Additional Ablations}
\paragraph{Effect of number of slots.}
We test the dependence of \ourmethod on the number of slots in the slot-attention module. \ourmethod is normally trained with 7 slots, consistent with previous OCL models. \cref{tab:appendix:effect_num_slots} additionally shows results for instance-aware object discovery task when GLASS is trained with 14 and 21 slots. As seen, increasing the number of slots leads to a decrease in mIoU$_i$ and mBO$_i$, indicating that the model cannot segment the objects correctly. This is because if the number of objects is larger than the number of objects in the scene, the scene is over-segmentated. The larger the number of slots, the more slots bind to object parts.

\input{tables/nips_tabs/ablation_num_slots}

\myparagraph{Effect of caption-generation module.}
\Cref{tab:appendix:effect_caption} shows an analysis of the effect of the caption generation module on the instance-aware object discovery task. \ourmethod uses a BLIP-2 \cite{li2023blip} model. We also test the performance against a more powerful captioning model, namely ShareGPT-4V \cite{chen2024sharegpt4v}, and a simple template-based pipeline. 

In the template-based pipeline, we first determine the empirical probability of each class appearing in the image using ground-truth class labels from the COCO dataset and the empirical probability of a certain number of objects appearing in a COCO image. Following this, we first sample the number of objects and then sample class labels from the learnt object occurrence distribution. After this, we populate the standard template ``A high-quality image of $<$obj(i)$>$, $<$obj(i+1)$>$ \ldots $<$obj(k)$>$; $<$obj(i)$>$ $<$obj(i+1)$>$ \ldots $<$object (k)$>$" using the class labels of the sampled objects.  
 \textbf{Note:} For this template-based pipeline, no input image is needed for caption generation. However, the dataset statistics in terms of the probability of occurrence of the objects and the probability of a certain number of objects in an image are required.  

\input{tables/nips_tabs/ablation_captions}

We find that our approach is not very sensitive to the particular choice of captioning model.
Interestingly, the template-based approach slightly outperforms both captioning models, showing that we largely need to ensure that the generated images possess the appropriate object occurrence statistics.

%% file: tables/nips_tabs/training_details.tex
\newcolumntype{Y}{>{\centering\arraybackslash}X}
\newcolumntype{L}{>{\raggedright\arraybackslash}X}
\begin{table}[b]
\centering

\smallskip
\scriptsize
\begin{tabularx}{\linewidth}{@{}lLc@{}}
\toprule
Module            &  Hyperparameter             &  Value\\ 
\midrule
General     &   Batch size       & 32  \\     
            &   Precision        & fp16    \\
            &   Learning rate: Phase 1    & 2e-5      \\
            &   Learning rate: Phase 2    & 4e-5    \\
            &   Optimizer        & Adam    \\
            &   Learning rate scheduler & Constant\\
\midrule
Encoder     &   Architecture             & DINOv2   \\
            &   Patch size       & 14       \\
            &   Backbone         & ViT-B    \\
            &   Embedding dimensions   & 768   \\
\midrule
Decoder-1     & Architecture               & Stable Diffusion\\
            & Model version         & 2.1         \\
\midrule
            Decoder-2 & Architecture & MLP \\
            & No.\ of layers & 3  \\
            & Hidden dimensions & 1536 \\
\bottomrule

\end{tabularx}
\vspace{-0.5em}
\caption{\textbf{Training details for GLASS and GLASS$^\dagger$.}}
\label{tab:appendix:training_detail}
\vspace{-0.5em}
\end{table}

%% file: tables/nips_tabs/comparision_fgari.tex
\begin{table}[tbp]
    \centering
    \scriptsize
    \smallskip    
    \begin{tabularx}{\columnwidth}
    {>{\hspace{-\tabcolsep}\raggedright\columncolor{white}[\tabcolsep][\tabcolsep]}XS[table-format=2.1]S[table-format=2.1]}
    
        \toprule
        Model & \multicolumn{2}{c}{FG-ARI (\%, $\uparrow)$} \\
        \cmidrule(lr){2-3}
        & VOC & COCO \\
        \midrule
        SA$^*$ \cite{locatello2020object} \tiny{NeurIPS'20} & 12.3 & 21.4 \\
        SLATE$^*$ \cite{singh2021illiterate} \tiny{ICLR'22} & 15.6 & 32.5 \\
        DINOSAUR-MLP \cite{seitzer2022bridging} \tiny{ICLR'23} & \best{24.6} & \best{40.6} \\ 
        DINOSAUR-Trans. \cite{seitzer2022bridging} \tiny{ICLR'23} & \secondbest{23.1} & 35.2 \\ 
        SPOT \cite{kakogeorgiou2024spot} \tiny{CVPR'24} & 20.9 & \secondbest{36.5} \\ 
        SlotDiffusion$^*$ \cite{wu2023slotdiffusion} \tiny{NeurIPS'23} & 17.8  & 37.2 \\
        StableLSD \cite{jiang2023object} \tiny{NeurIPS'23} & 8.7 & 28.9 \\ 
        \rowcolor{gray!15} \ourmethodDagger (ours) & 21.3 &  32.5 \\ 
        \rowcolor{gray!15} GLASS (ours) & 22.5 & 34.1 \\ 
        \bottomrule
    \end{tabularx}
    \vspace{-0.75em} 
    \caption{\textbf{Comparison between OCL methods for the FG-ARI metric.} Here, our method is \emph{only} behind DINOSAUR on the VOC dataset and performs close to SPOT on the COCO dataset. Please note that the FG-ARI metric is unreliable as it does not take into account the shape of the predicted mask and also ignores the background, making it unsuitable for object discovery as seen by the results in \cref{fig:appendix:OD}. * denotes numbers taken from \cite{kakogeorgiou2024spot}.}
    \label{tab:supp:fgari}
    \vspace{-0.5em}
\end{table}

%% file: figures/sam2_appx.tex
\begin{figure}
    \scriptsize
    \setlength{\tabcolsep}{1pt}
    \centering
    \begin{tabular}{@{}
    *{3}{x{70}}
    @{}}
    Input & SAMv2 (Under-segmentation) & SAMv2 (Over-segmentation) \\
    \includegraphics[width=0.14\textwidth]{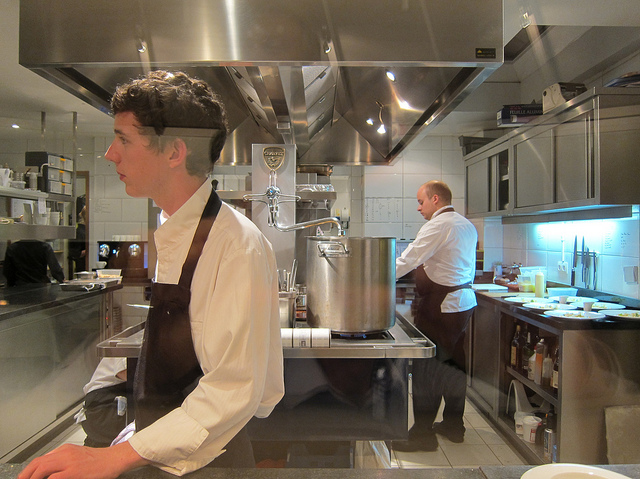}
    & \includegraphics[width=0.14\textwidth]{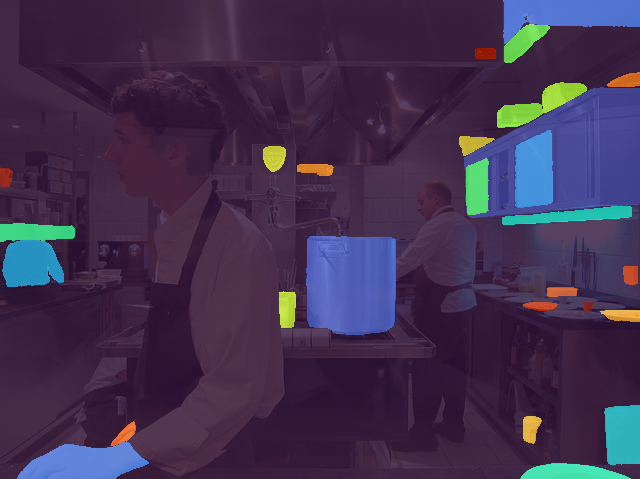}
    & \includegraphics[width=0.14\textwidth]{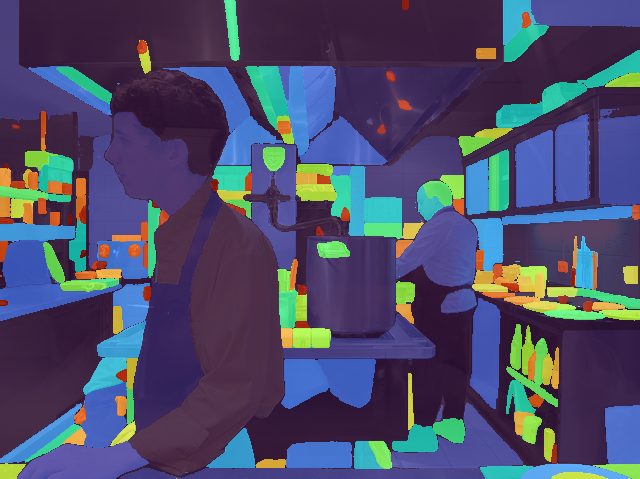} \\
    \includegraphics[width=0.14\textwidth]{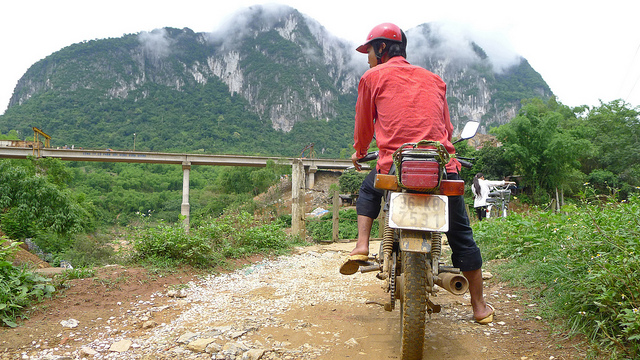}
    & \includegraphics[width=0.14\textwidth]{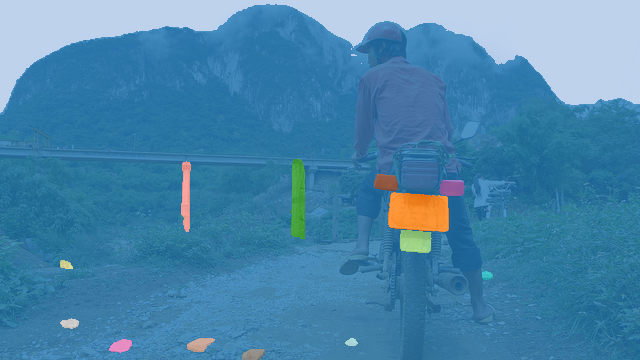}
    & \includegraphics[width=0.14\textwidth]{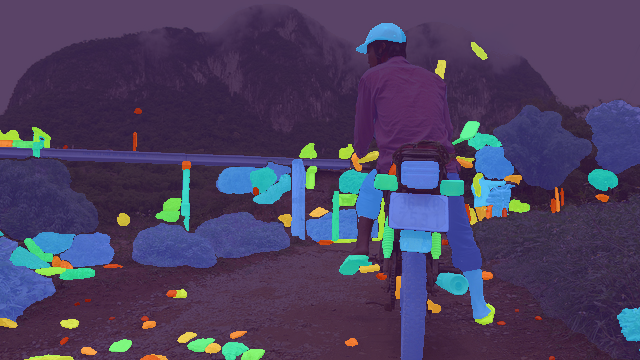} \\
    \includegraphics[width=0.14\textwidth]{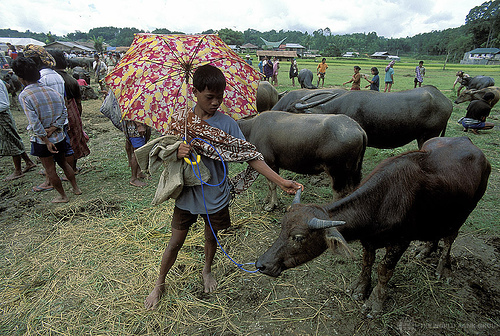}
    & \includegraphics[width=0.14\textwidth]{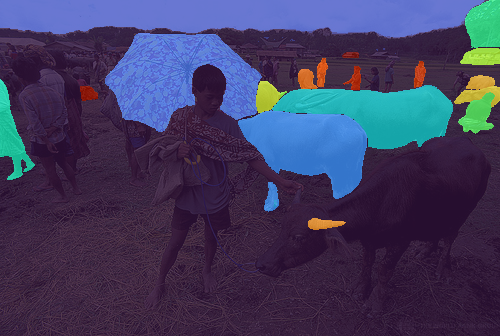}
    & \includegraphics[width=0.14\textwidth]{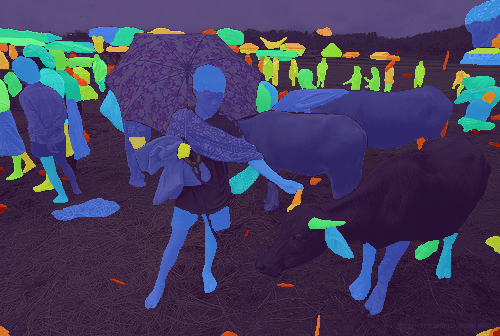} \\

    \end{tabular}
    \vspace{-0.5em}
    \caption{\textbf{Issue with SAMv2 masks as guiding signal.} Automatic segmentation using SAMv2 is very sensitive to the choice of hyperparameters, which makes it suffer from over- or under-segmentation issues. Hence, automatic SAMv2 masks are not ideal as guidance signal.} 
    \label{fig:appendix:sam_failure}
    \vspace{-0.5em}
\end{figure}

%% file: figures/compo_gen_stablelsd.tex
\begin{figure}
    \centering
    \footnotesize
    \setlength{\tabcolsep}{1pt}
    \begin{tabular}{@{}
    *{3}{x{60}} 
    @{}}
    Add item  & Original image & Edited image  \\
    \includegraphics[width=0.10\textwidth]{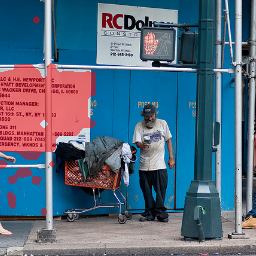}
    & \includegraphics[width=0.10\textwidth]{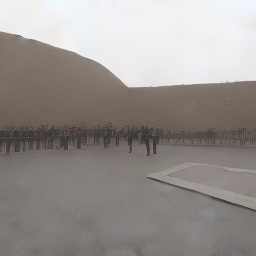}
    & \includegraphics[width=0.10\textwidth]{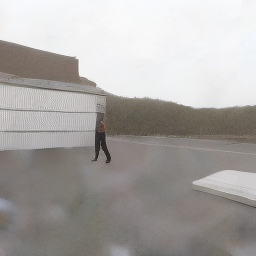} \\
    \includegraphics[width=0.10\textwidth]{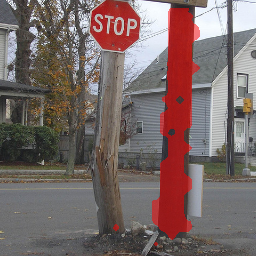}
    & \includegraphics[width=0.10\textwidth]{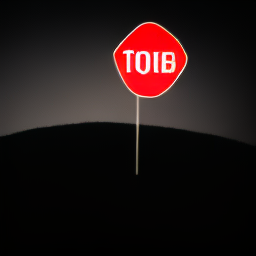}
    & \includegraphics[width=0.10\textwidth]{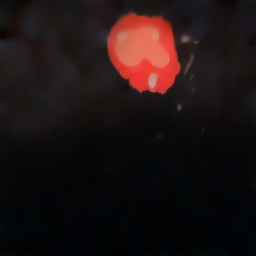} \\
    Remove item  & Original image & Edited image \\
    \includegraphics[width=0.10\textwidth]{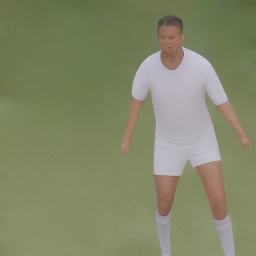}
    & \includegraphics[width=0.10\textwidth]{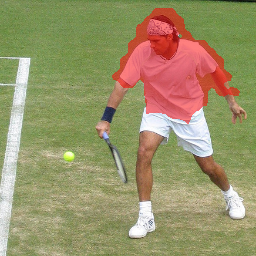}
    & \includegraphics[width=0.10\textwidth]{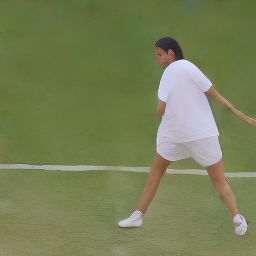} \\
    
    \includegraphics[width=0.10\textwidth]{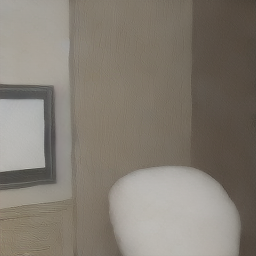}
    & \includegraphics[width=0.10\textwidth]{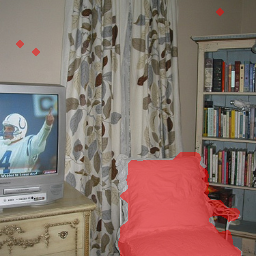}
    & \includegraphics[width=0.10\textwidth]{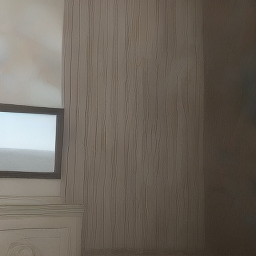} \\
     \includegraphics[width=0.10\textwidth]{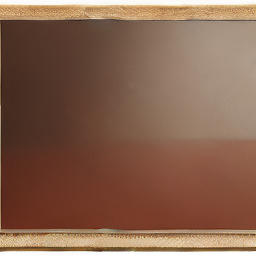}
    & \includegraphics[width=0.10\textwidth]{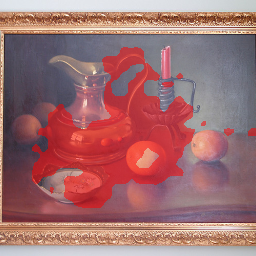}
    & \includegraphics[width=0.10\textwidth]{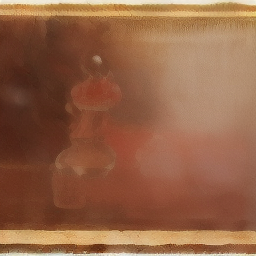} \\
    \end{tabular}
    \vspace{-.5em}
    \caption{\textbf{Compositional generation with StableLSD \cite{jiang2023object}.} The quality of the image reconstruction is rather poor for StableLSD. Moreover, the approach does not always remove the annotated item from the image \emph{(row 3 and 5)}. Adding of an item to a new scene also results in failure as is the case with \emph{(row 2)}. There is some compositionality, which is exhibited by \emph{(rows 1 and 4)}, but the quality of image reconstruction is poor and the edited image is not faithful to the original image.}
    \label{fig:appendix:comp_gen_lsd}
    \vspace{-0.5em}
\end{figure}

%% file: figures/compo_gen_cmp_comp_diffusion.tex
\begin{figure}
    \centering
    \footnotesize
    \setlength{\tabcolsep}{1pt}
    \begin{tabular}{@{}*{4}{x{60}}@{}}
    Src. Image & Dst. Image & \ourmethod & Composable Diffusion \\
    \includegraphics[width=0.110\textwidth]{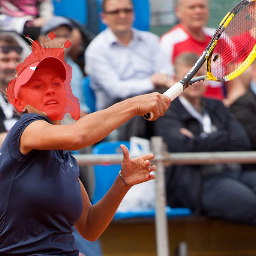} &
    \includegraphics[width=0.110\textwidth]{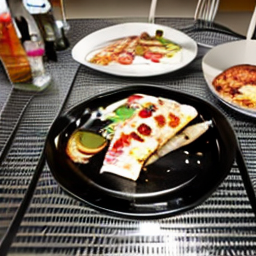} &
    \includegraphics[width=0.110\textwidth]{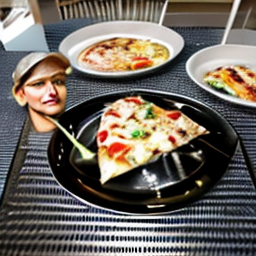} &
    \includegraphics[width=0.110\textwidth]{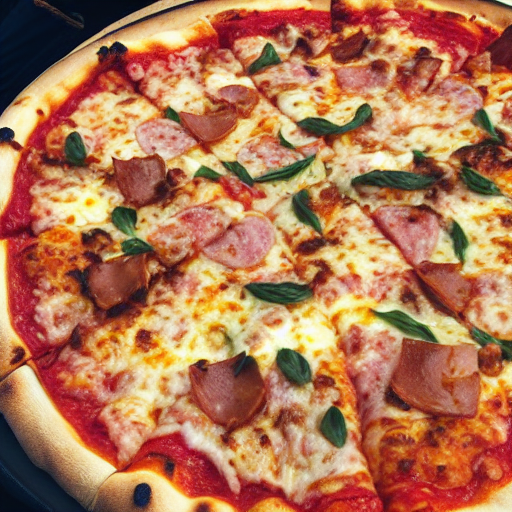} \\
    \includegraphics[width=0.110\textwidth]{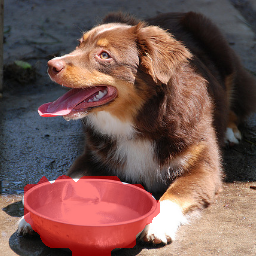} &
    \includegraphics[width=0.110\textwidth]{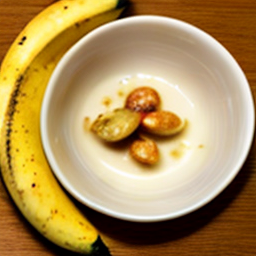} &
    \includegraphics[width=0.110\textwidth]{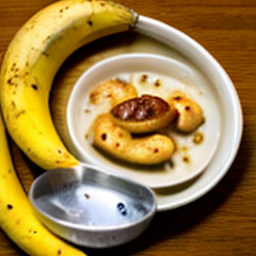} &
    \includegraphics[width=0.110\textwidth]{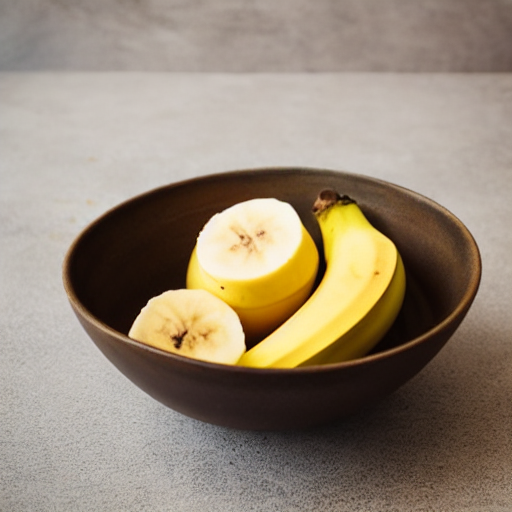} \\
    \includegraphics[width=0.110\textwidth]{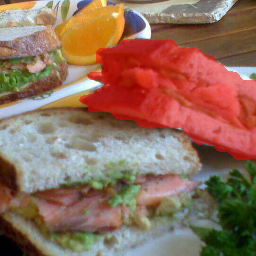} &
    \includegraphics[width=0.110\textwidth]{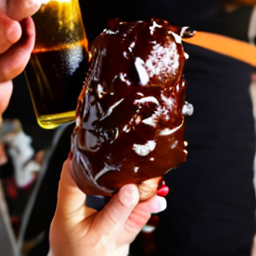} &
    \includegraphics[width=0.110\textwidth]{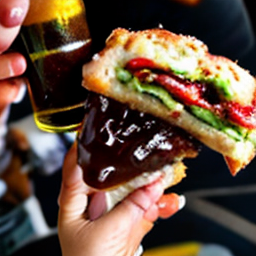} &
    \includegraphics[width=0.110\textwidth]{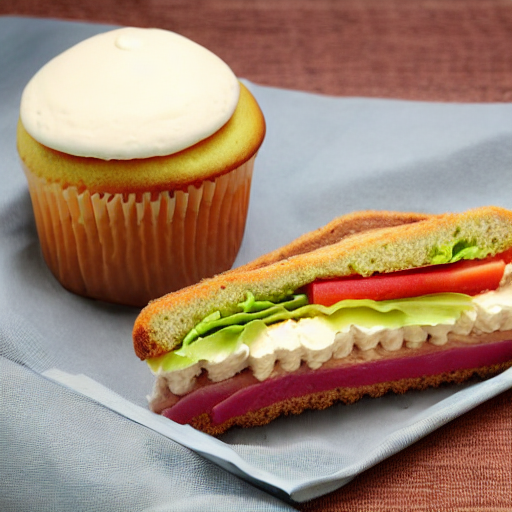} \\
    \includegraphics[width=0.110\textwidth]{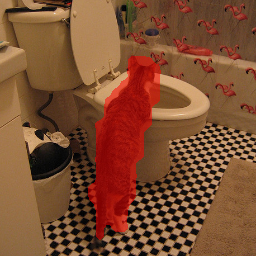} &
    \includegraphics[width=0.110\textwidth]{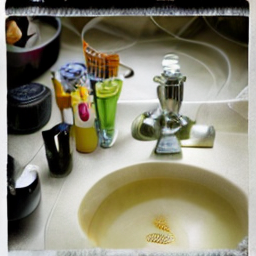} &
    \includegraphics[width=0.110\textwidth]{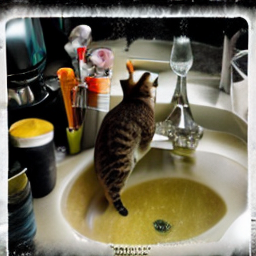} &
    \includegraphics[width=0.110\textwidth]{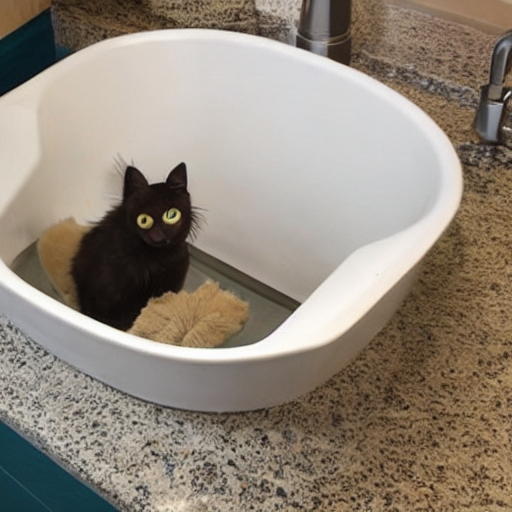} \\
    \end{tabular}
    \vspace{-1em}
    \caption{\textbf{Qualitative comparison for concept composition with Composable Diffusion \cite{liu2022compositional}.} \ourmethod is able to extract and compose concepts/slots from a source (Src.) and transfer them to a destination image (Dst.) image to generate an image containing concepts/slots from both images. We compare this to text-based compositional models \cite{liu2022compositional}, where we compose the labels from the two images, \emph{(row-1)} ``Pizza and Person", \emph{(row-2)} ``BOWL AND BOWL AND BANANA", \emph{(row-3)}, ``CUP AND CAKE AND SANDWICH", and \emph{(row-4)} ``CAT AND SINK" to generate an image containing both concepts. As seen, our results have a higher fidelity, more realism, and are more faithful to the concepts given.}
    \label{fig:appendix:comp_gen_cmp_comp_diffusion}
    \vspace{-1em}
\end{figure}

%% file: figures/od_qualitative_appx.tex
\definecolor{mymethodcolor}{RGB}{101, 143, 255}

\begin{figure*}
    \scriptsize
    \setlength{\tabcolsep}{1pt}
    \centering
    \begin{tabular}{@{}
    *{6}{x{70}}
    @{}}
    Input & DINOSAUR \cite{seitzer2022bridging} & StableLSD \cite{jiang2023object} & SPOT \cite{kakogeorgiou2024spot} & \textbf{GLASS$^{\dagger}$ (ours)} & \textbf{GLASS (ours)} \\
    \includegraphics[width=0.14\textwidth]{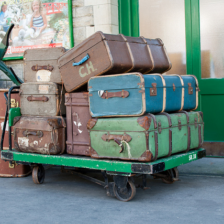} 
    &\includegraphics[width=0.14\textwidth]{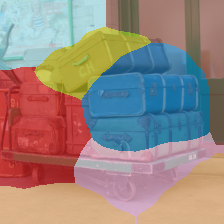}
    &\includegraphics[width=0.14\textwidth]{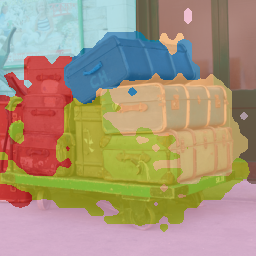}
    &\includegraphics[width=0.14\textwidth]{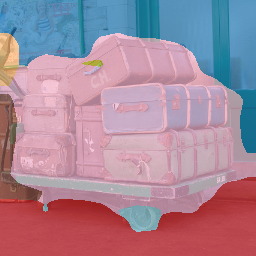}
    &\includegraphics[width=0.14\textwidth]{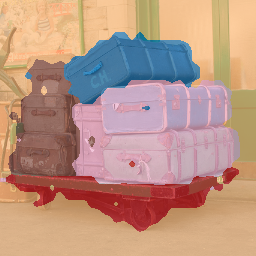}
    &\includegraphics[width=0.14\textwidth]{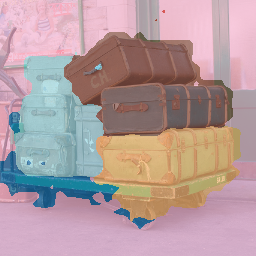}
    \\
    \includegraphics[width=0.14\textwidth]{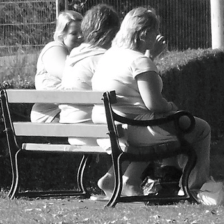} 
    &\includegraphics[width=0.14\textwidth]{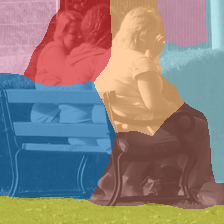}
    &\includegraphics[width=0.14\textwidth]{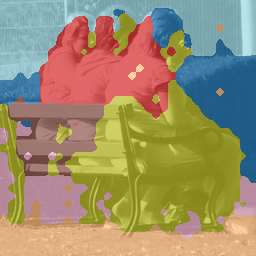}
    &\includegraphics[width=0.14\textwidth]{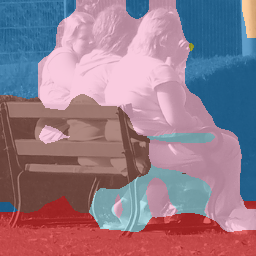}
    &\includegraphics[width=0.14\textwidth]{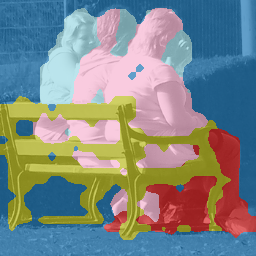}
    &\includegraphics[width=0.14\textwidth]{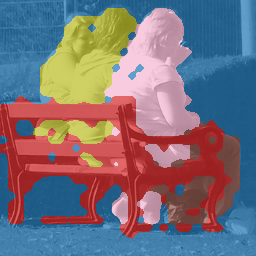}
     \\
     \scalebox{-1}[1]{\includegraphics[width=0.14\textwidth]{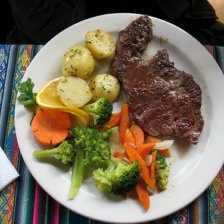}}
    & \scalebox{-1}[1]{\includegraphics[width=0.14\textwidth]{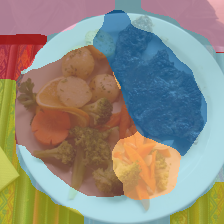}}
    &\includegraphics[width=0.14\textwidth]{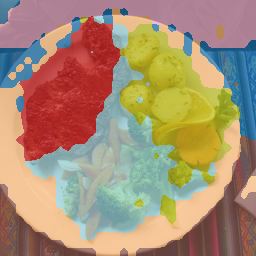}
    &\includegraphics[width=0.14\textwidth]{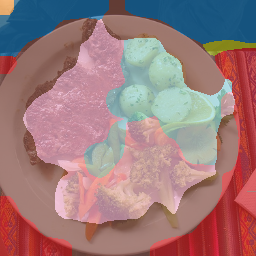}
    &\includegraphics[width=0.14\textwidth]{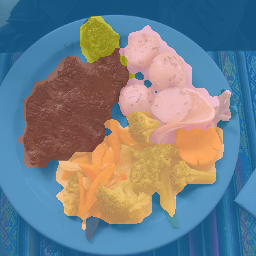}
    &\includegraphics[width=0.14\textwidth]{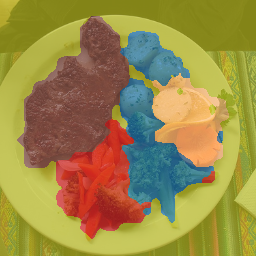}
    \\
    \scalebox{-1}[1]{
    \includegraphics[width=0.14\textwidth]{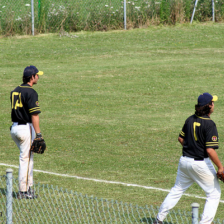}}
    &\scalebox{-1}[1]{\includegraphics[width=0.14\textwidth]{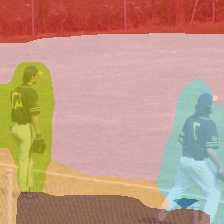}} &\includegraphics[width=0.14\textwidth]{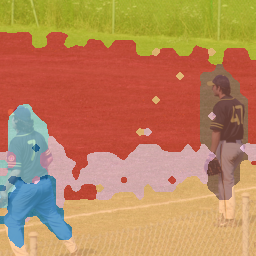}
    &\includegraphics[width=0.14\textwidth]{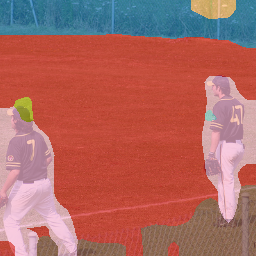}
    &\includegraphics[width=0.14\textwidth]{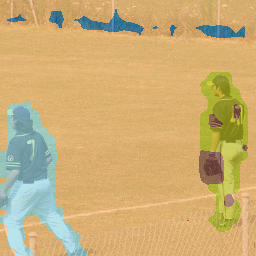}
    &\includegraphics[width=0.14\textwidth]{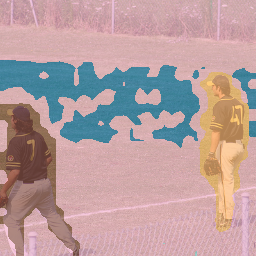}
    \\
    \includegraphics[width=0.14\textwidth]{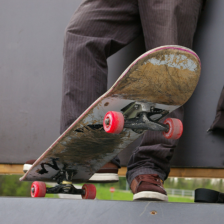} 
    &\includegraphics[width=0.14\textwidth]{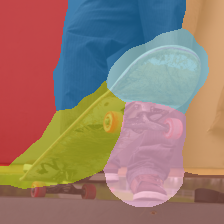} &\includegraphics[width=0.14\textwidth]{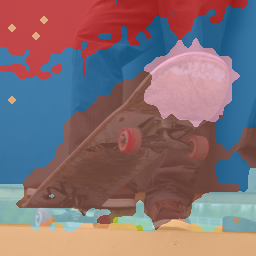}
    &\includegraphics[width=0.14\textwidth]{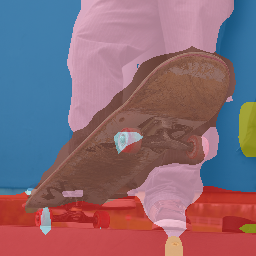}
    &\includegraphics[width=0.14\textwidth]{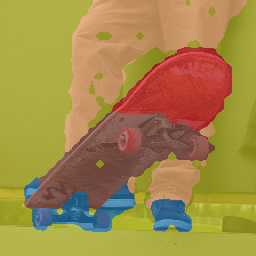}
    &\includegraphics[width=0.14\textwidth]{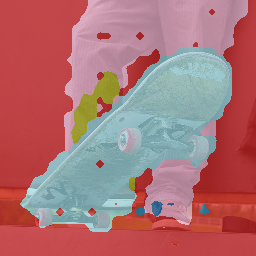}
    \\
    \scalebox{-1}[1]{
    \includegraphics[width=0.14\textwidth]{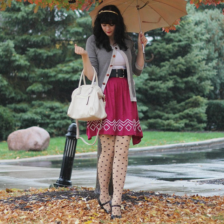}}
    &
    \scalebox{-1}[1]{\includegraphics[width=0.14\textwidth]{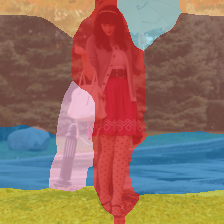}} &\includegraphics[width=0.14\textwidth]{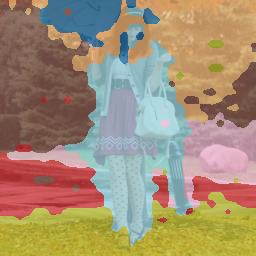}
    &\includegraphics[width=0.14\textwidth]{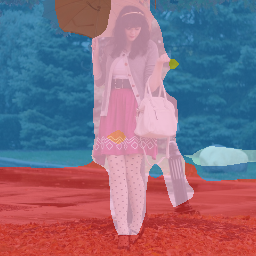}
    &\includegraphics[width=0.14\textwidth]{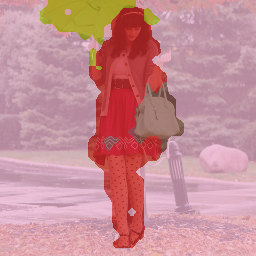}
    &\includegraphics[width=0.14\textwidth]{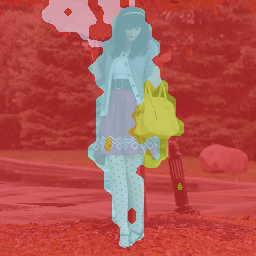}
    \\

    \scalebox{-1}[1]{
    \includegraphics[width=0.14\textwidth]{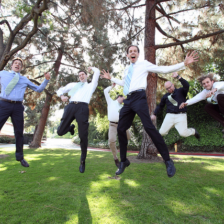}}
    &
    \scalebox{-1}[1]{
    \includegraphics[width=0.14\textwidth]{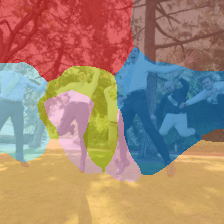}} 
    &\includegraphics[width=0.14\textwidth]{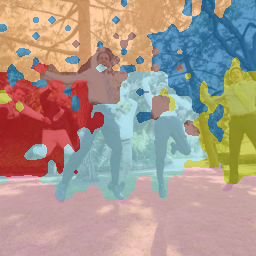}
    &\includegraphics[width=0.14\textwidth]{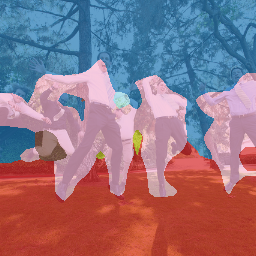}
    &\includegraphics[width=0.14\textwidth]{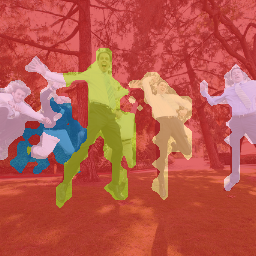}
    &\includegraphics[width=0.14\textwidth]{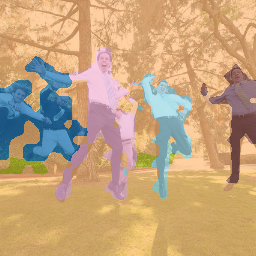}
    \\
    \includegraphics[width=0.14\textwidth]{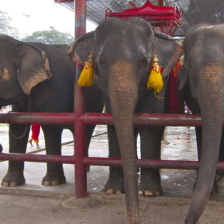} 
    &\includegraphics[width=0.14\textwidth]{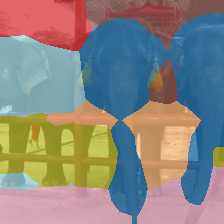}
    &\includegraphics[width=0.14\textwidth]{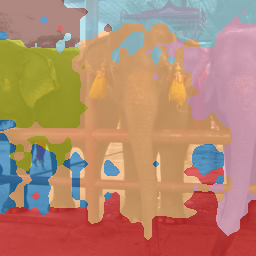}
    &\includegraphics[width=0.14\textwidth]{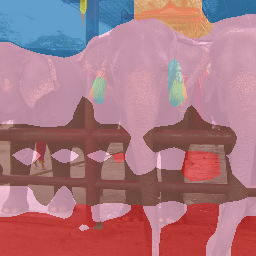}
    &\includegraphics[width=0.14\textwidth]{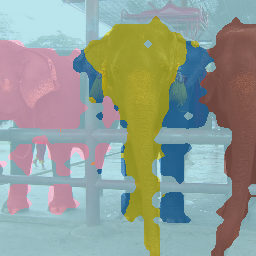}
    &\includegraphics[width=0.14\textwidth]{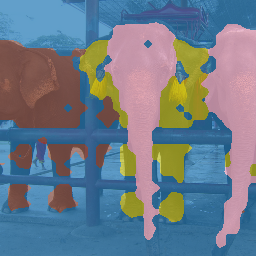}
    \end{tabular}
    \vspace{-1em}
    \caption{\textbf{Qualitative comparison for object discovery.} \ourmethod and \ourmethodDagger can decompose an image at the object level and do not split an object into its parts or group objects belonging to the same class. Also, our approach yields cleaner boundaries for the foreground objects compared to DINOSAUR \cite{seitzer2022bridging}, StableLSD \cite{jiang2023object}, and SPOT \cite{kakogeorgiou2024spot}.} 
    \label{fig:appendix:OD}
\end{figure*}

%% file: figures/fid_qual_appx.tex
\begin{figure*}[p]
    \centering
    \footnotesize
    \setlength{\tabcolsep}{1pt}
    \begin{tabular}{@{}*{3}{x{75}}|*{3}{x{75}}@{}}
    Original & StableLSD \cite{jiang2023object} & \ourmethod & Original & StableLSD \cite{jiang2023object} & \ourmethod \\
    \includegraphics[width=0.15\textwidth]{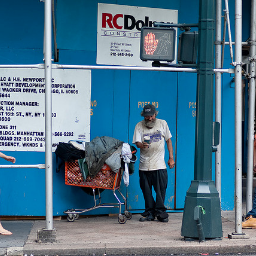} &
    \includegraphics[width=0.15\textwidth]{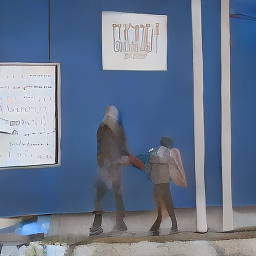} &
    \includegraphics[width=0.15\textwidth]{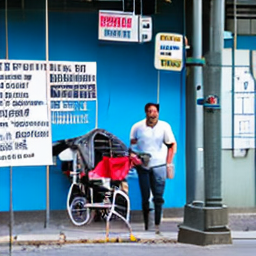} &
    \includegraphics[width=0.15\textwidth]{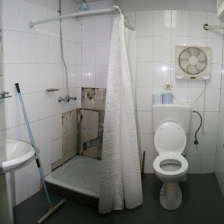} &
    \includegraphics[width=0.15\textwidth]{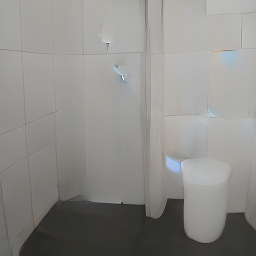} &
    \includegraphics[width=0.15\textwidth]{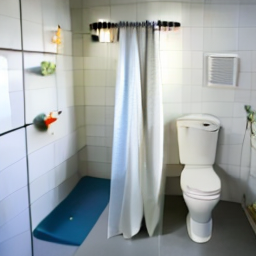} \\
    \includegraphics[width=0.15\textwidth]{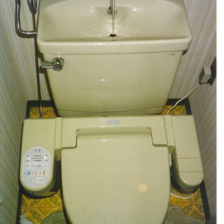} &
    \includegraphics[width=0.15\textwidth]{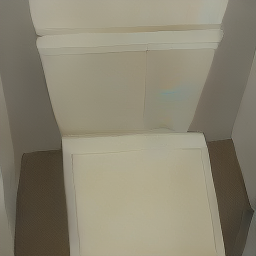} &
    \includegraphics[width=0.15\textwidth]{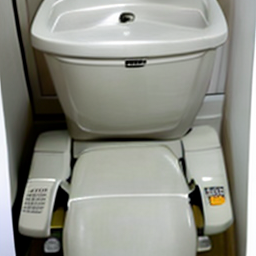} &
    \includegraphics[width=0.15\textwidth]{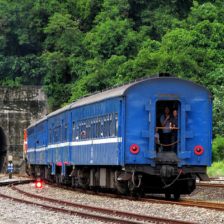} &
    \includegraphics[width=0.15\textwidth]{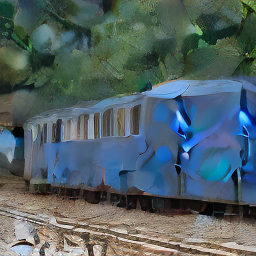} &
    \includegraphics[width=0.15\textwidth]{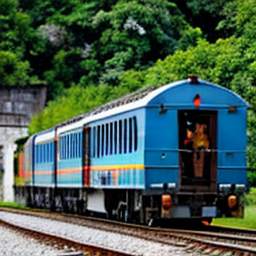} \\
    \includegraphics[width=0.15\textwidth]{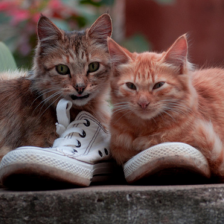} &
    \includegraphics[width=0.15\textwidth]{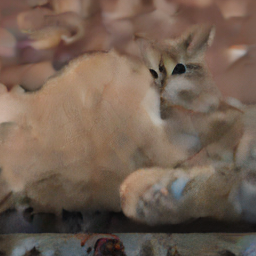} &
    \scalebox{-1}[1]{
    \includegraphics[width=0.15\textwidth]{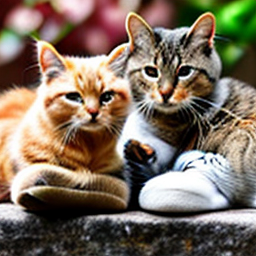}}s &
    \includegraphics[width=0.15\textwidth]{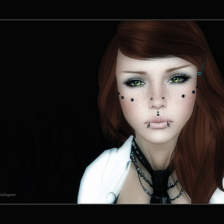} &
    \includegraphics[width=0.15\textwidth]{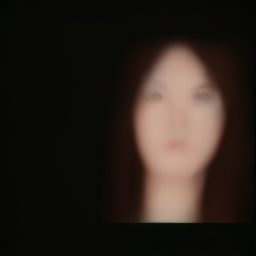} &
    \includegraphics[width=0.15\textwidth]{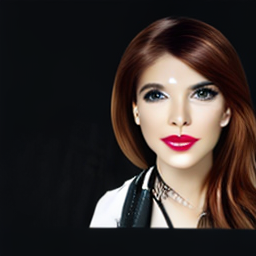} \\
    \includegraphics[width=0.15\textwidth]{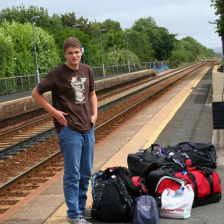} &
    \includegraphics[width=0.15\textwidth]{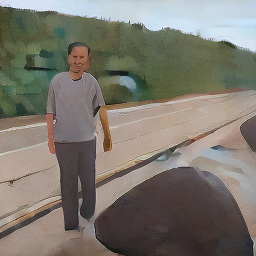} &
    \includegraphics[width=0.15\textwidth]{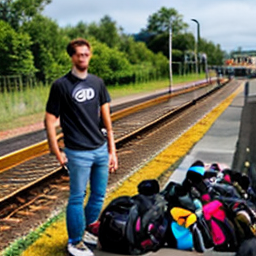} &
    \includegraphics[width=0.15\textwidth]{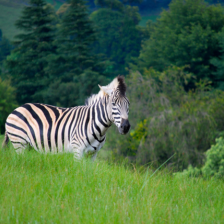} &
    \includegraphics[width=0.15\textwidth]{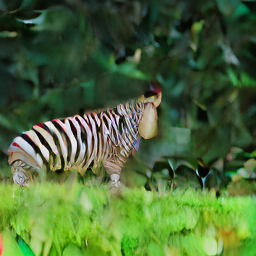} &
    \includegraphics[width=0.15\textwidth]{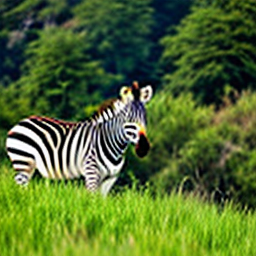} \\
    \includegraphics[width=0.15\textwidth]{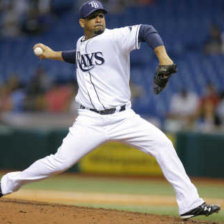} &
    \includegraphics[width=0.15\textwidth]{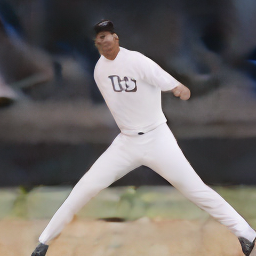} &
    \includegraphics[width=0.15\textwidth]{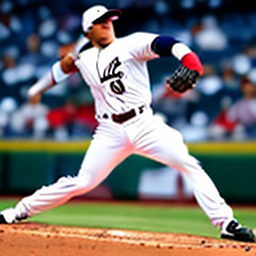} &
    \includegraphics[width=0.15\textwidth]{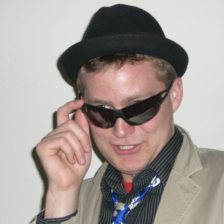} &
    \includegraphics[width=0.15\textwidth]{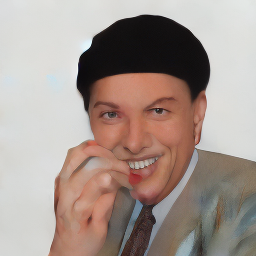} &
    \includegraphics[width=0.15\textwidth]{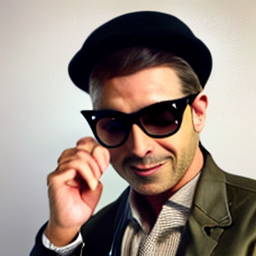} \\
    \includegraphics[width=0.15\textwidth]{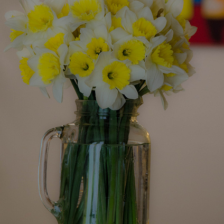} &
    \includegraphics[width=0.15\textwidth]{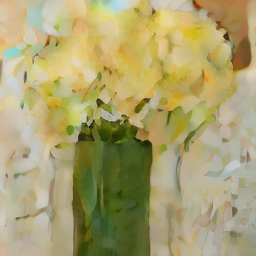} &
    \includegraphics[width=0.15\textwidth]{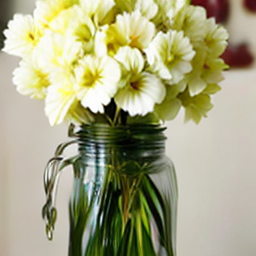} &
    \includegraphics[width=0.15\textwidth]{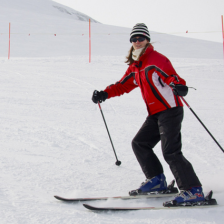} &
    \includegraphics[width=0.15\textwidth]{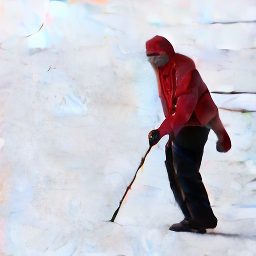} &
    \includegraphics[width=0.15\textwidth]{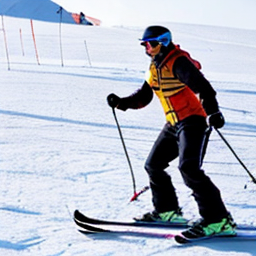} \\
    \includegraphics[width=0.15\textwidth]{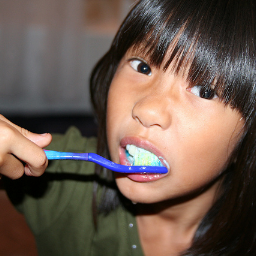} &
    \includegraphics[width=0.15\textwidth]{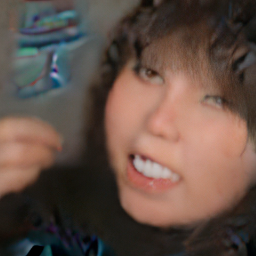} &
    \includegraphics[width=0.15\textwidth]{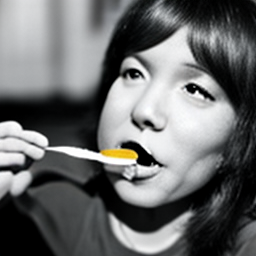} &
    \includegraphics[width=0.15\textwidth]{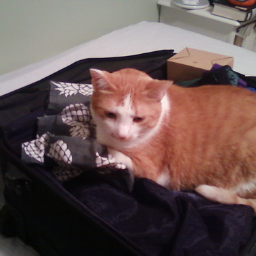} &
    \includegraphics[width=0.15\textwidth]{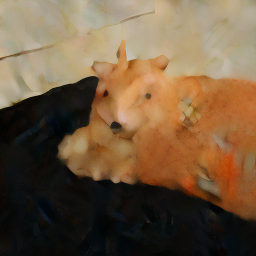} &
    \includegraphics[width=0.15\textwidth]{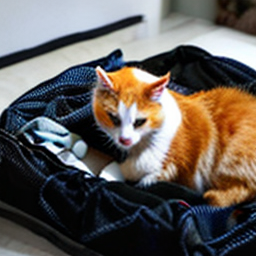} \\
    \end{tabular}
    \vspace{-1em}
    \caption{\textbf{Qualitative comparison for conditional image generation.} \ourmethod and \ourmethodDagger not only learn to decompose scenes meaningfully, but the learned slot can reconstruct the input scene more faithfully and with higher fidelity than StableLSD \cite{jiang2023object}.}
    \label{fig:appendix:CG}
    \vspace{-1em}
\end{figure*}

%% file: figures/compo_gen_remove_appx.tex
\begin{figure*}[p]
    \centering
    \footnotesize
    \setlength{\tabcolsep}{1pt}
    \begin{tabular}{@{}
    *{3}{x{75}}|*{3}{x{75}} 
    @{}}
    Remove Item  & Original image & Edited image & Remove Item  & Original image & Edited image \\
    \includegraphics[width=0.15\textwidth]{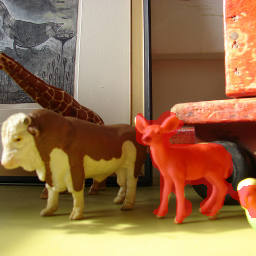}
    & \includegraphics[width=0.15\textwidth]{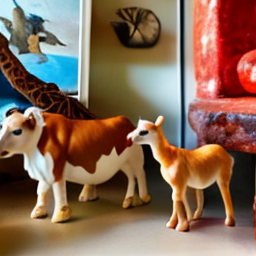} 
    & \includegraphics[width=0.15\textwidth]{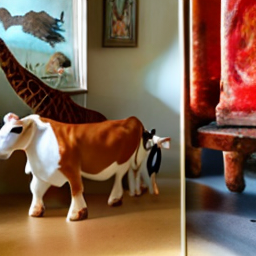} 
    & \includegraphics[width=0.15\textwidth]{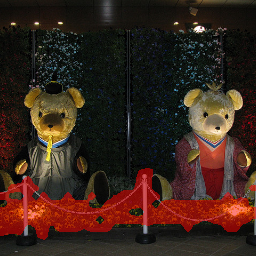}
    & \includegraphics[width=0.15\textwidth]{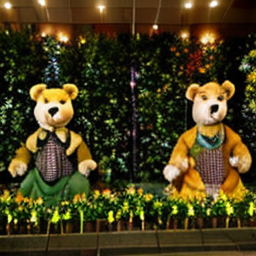} 
    & \includegraphics[width=0.15\textwidth]{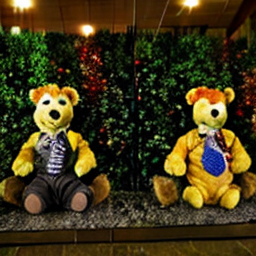} \\
     \includegraphics[width=0.15\textwidth]{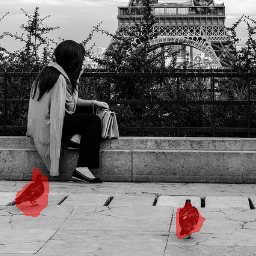}
    & \includegraphics[width=0.15\textwidth]{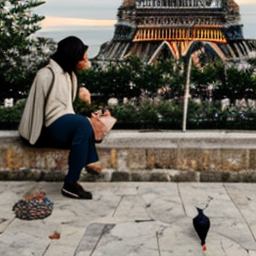}
    & \includegraphics[width=0.15\textwidth]{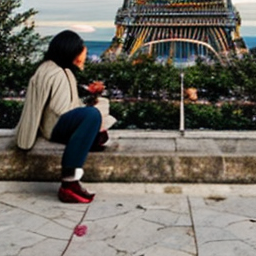}
    & \includegraphics[width=0.15\textwidth]{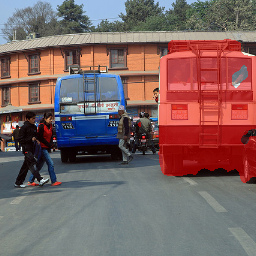}
    & \includegraphics[width=0.15\textwidth]{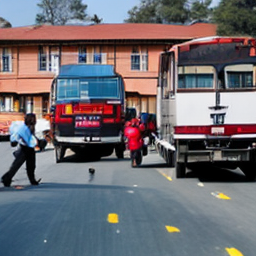} 
    & \includegraphics[width=0.15\textwidth]{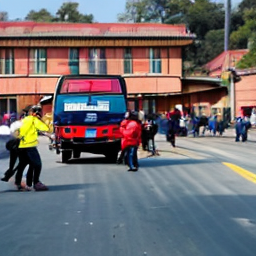} \\
    \includegraphics[width=0.15\textwidth]{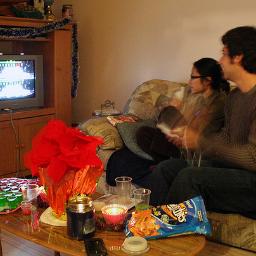}
    & \includegraphics[width=0.15\textwidth]{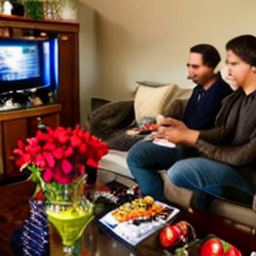} 
    & \includegraphics[width=0.15\textwidth]{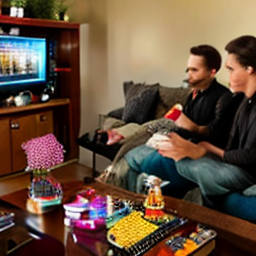} 
    & \includegraphics[width=0.15\textwidth]{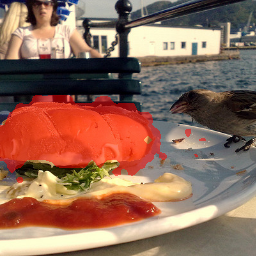}
    & \includegraphics[width=0.15\textwidth]{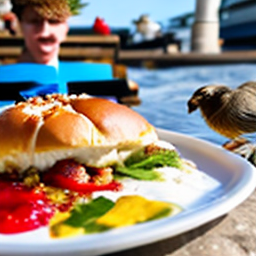} 
    & \includegraphics[width=0.15\textwidth]{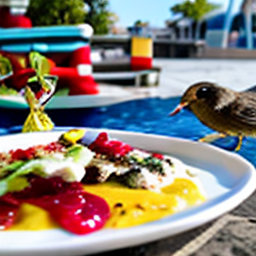} \\
    \includegraphics[width=0.15\textwidth]{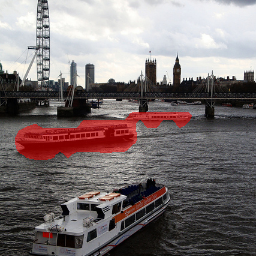}
    & \includegraphics[width=0.15\textwidth]{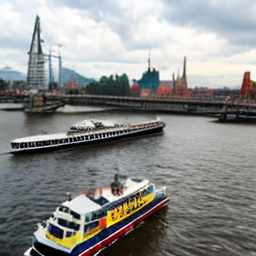}
    & \includegraphics[width=0.15\textwidth]{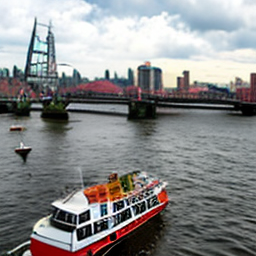}
    & \includegraphics[width=0.15\textwidth]{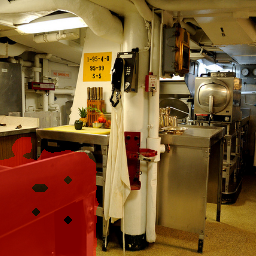}
    & \includegraphics[width=0.15\textwidth]{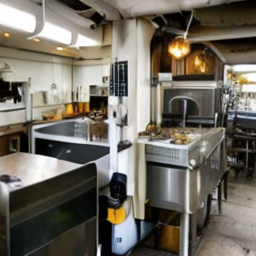}
    & \includegraphics[width=0.15\textwidth]{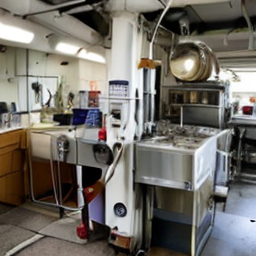} \\
    \includegraphics[width=0.15\textwidth]{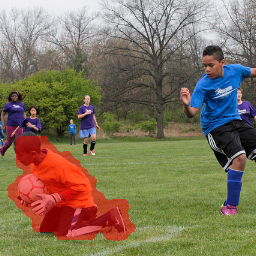}
    & \includegraphics[width=0.15\textwidth]{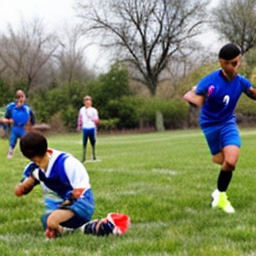}
    & \includegraphics[width=0.15\textwidth]{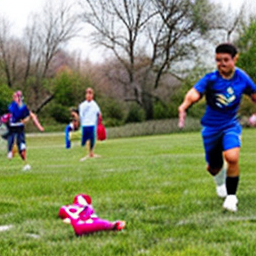}
    & \includegraphics[width=0.15\textwidth]{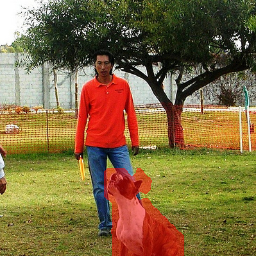}
    & \includegraphics[width=0.15\textwidth]{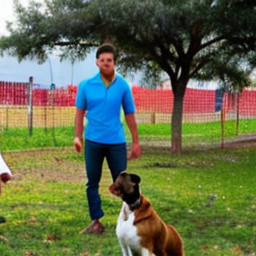}
    & \includegraphics[width=0.15\textwidth]{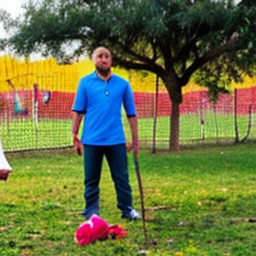} \\
    \includegraphics[width=0.15\textwidth]{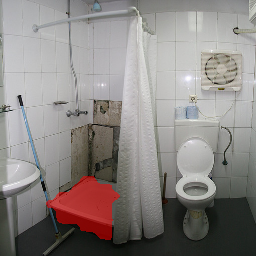}
    & \includegraphics[width=0.15\textwidth]{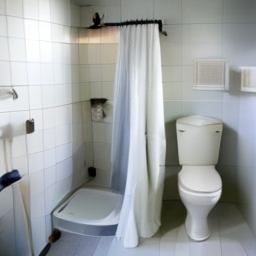}
    & \includegraphics[width=0.15\textwidth]{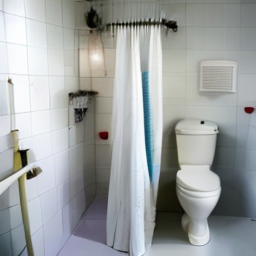}
    & \includegraphics[width=0.15\textwidth]{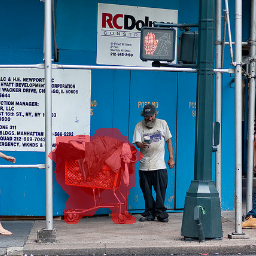}
    & \includegraphics[width=0.15\textwidth]{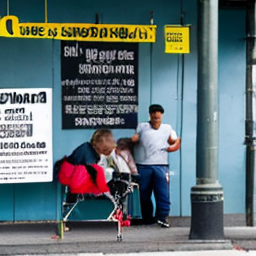}
    & \includegraphics[width=0.15\textwidth]{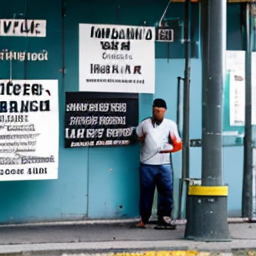} \\
    
    \end{tabular}
    \vspace{-.5em}
    \caption{\textbf{Compositional generation.} \ourmethod enables the compositional image generation of real-world complex scenes. Here, the masked object \emph{(in red)} is the slot to be removed from the original image. The original image is the reconstructed image from the slots of the input image.}
    \label{fig:appendix:comp_gen_remove}
    \vspace{-1em}
\end{figure*}

%% file: figures/compo_gen_add_appx.tex
\begin{figure*}[p]
    \centering
    \footnotesize
    \setlength{\tabcolsep}{1pt}
    \begin{tabular}{@{}
    *{3}{x{75}}|*{3}{x{75}} 
    @{}}
    Added item  & Original image & Edited image & Added item  & Original image & Edited image \\
    \includegraphics[width=0.15\textwidth]{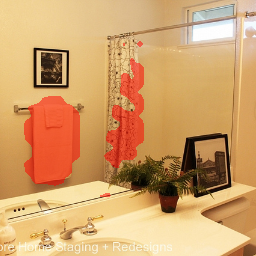}
    & \includegraphics[width=0.15\textwidth]{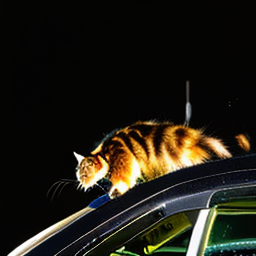}
    & \includegraphics[width=0.15\textwidth]{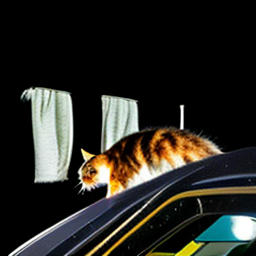}
    & \includegraphics[width=0.15\textwidth]{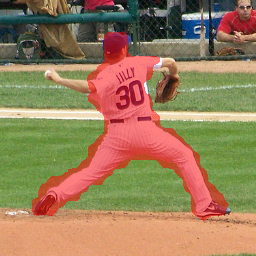}
    & \includegraphics[width=0.15\textwidth]{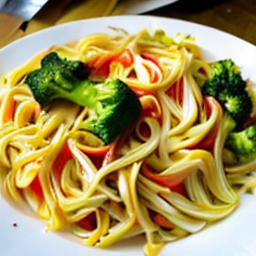}
    & \includegraphics[width=0.15\textwidth]{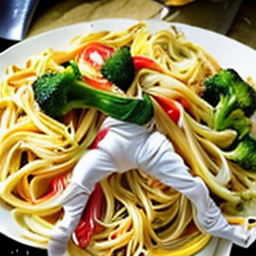} \\
    \includegraphics[width=0.15\textwidth]{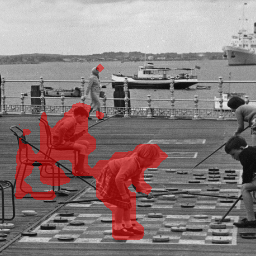}
    & \includegraphics[width=0.15\textwidth]{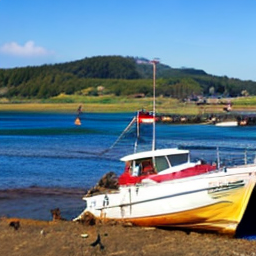}
    & \includegraphics[width=0.15\textwidth]{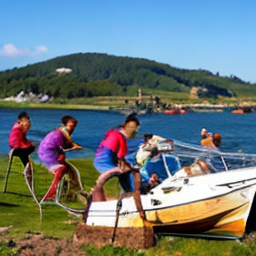}
    & \includegraphics[width=0.15\textwidth]{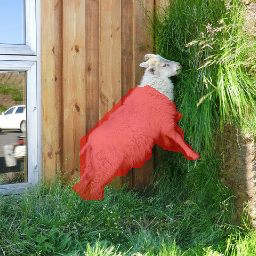}
    & \includegraphics[width=0.15\textwidth]{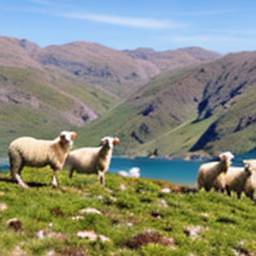}
    & \includegraphics[width=0.15\textwidth]{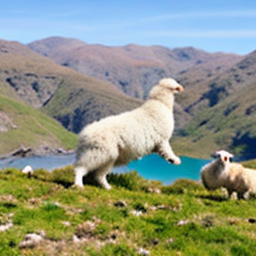} \\
    \includegraphics[width=0.15\textwidth]{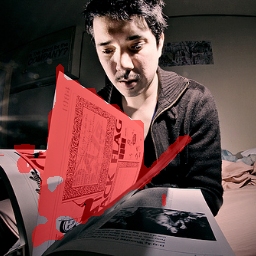}
    & \includegraphics[width=0.15\textwidth]{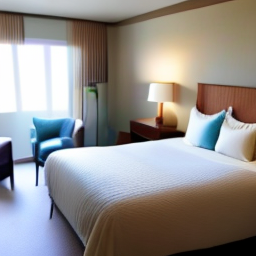}
    & \includegraphics[width=0.15\textwidth]{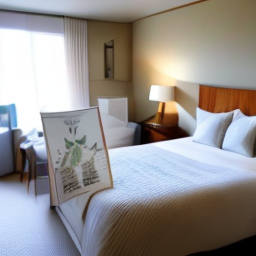}
    & \includegraphics[width=0.15\textwidth]{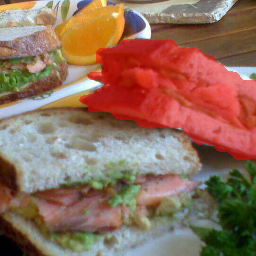}
    & \includegraphics[width=0.15\textwidth]{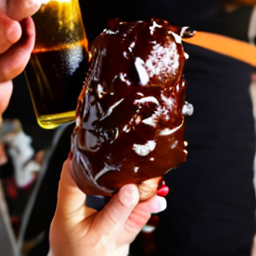}
    & \includegraphics[width=0.15\textwidth]{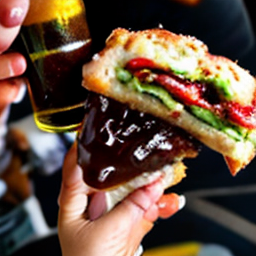} \\
     \includegraphics[width=0.15\textwidth]{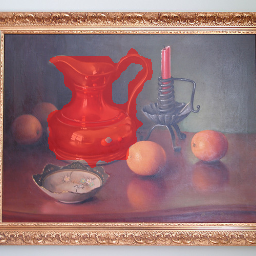}
    & \includegraphics[width=0.15\textwidth]{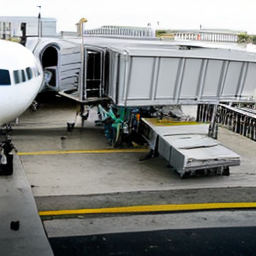}
    & \includegraphics[width=0.15\textwidth]{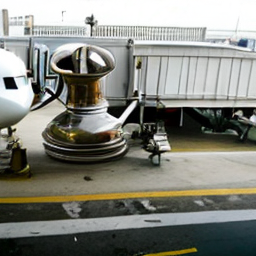}
    & \includegraphics[width=0.15\textwidth]{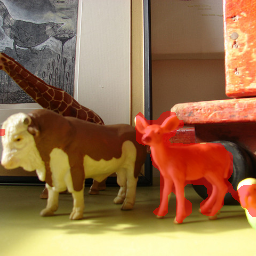}
    & \includegraphics[width=0.15\textwidth]{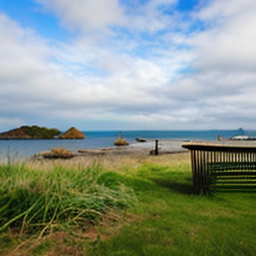}
    & \includegraphics[width=0.15\textwidth]{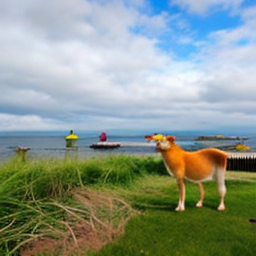} \\   
    \includegraphics[width=0.15\textwidth]{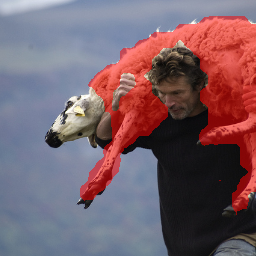}
    & \includegraphics[width=0.15\textwidth]{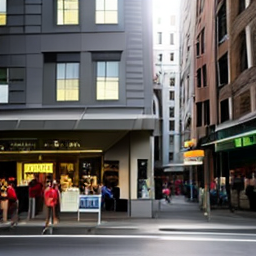}
    & \includegraphics[width=0.15\textwidth]{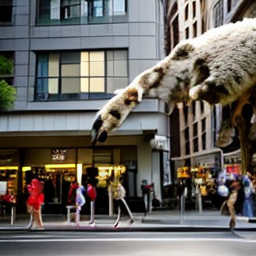}
    & \includegraphics[width=0.15\textwidth]{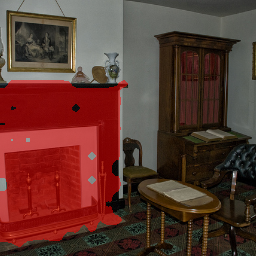}
    & \includegraphics[width=0.15\textwidth]{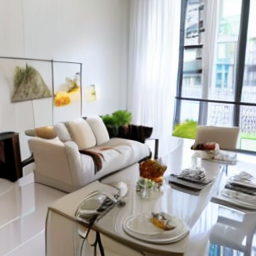}
    & \includegraphics[width=0.15\textwidth]{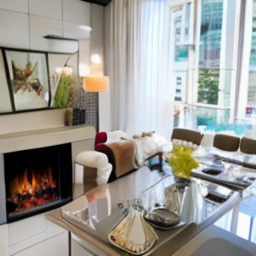} \\
    \includegraphics[width=0.15\textwidth]{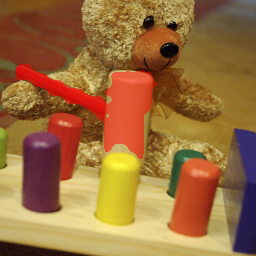}
    & \includegraphics[width=0.15\textwidth]{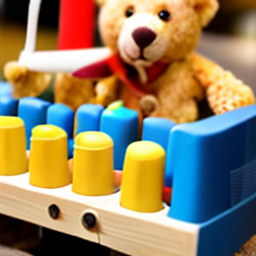}
    & \includegraphics[width=0.15\textwidth]{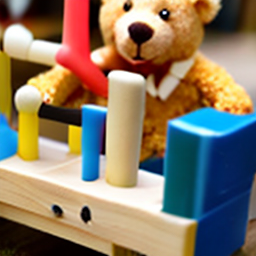}
    & \includegraphics[width=0.15\textwidth]{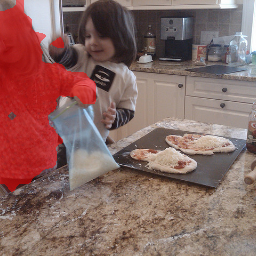}
    & \includegraphics[width=0.15\textwidth]{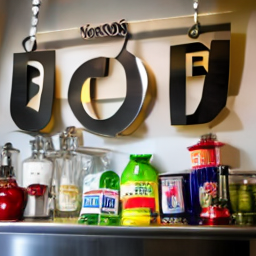}
    & \includegraphics[width=0.15\textwidth]{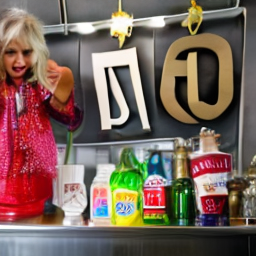} \\
    \includegraphics[width=0.15\textwidth]{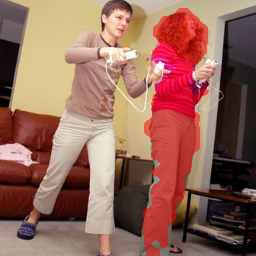}
    & \includegraphics[width=0.15\textwidth]{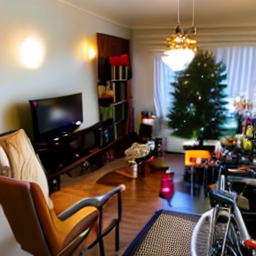}
    & \includegraphics[width=0.15\textwidth]{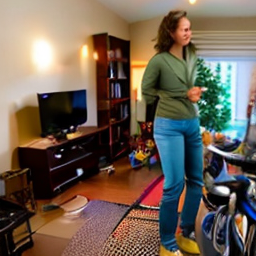}
    & \includegraphics[width=0.15\textwidth]{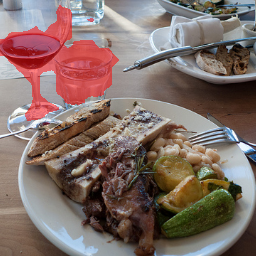}
    & \includegraphics[width=0.15\textwidth]{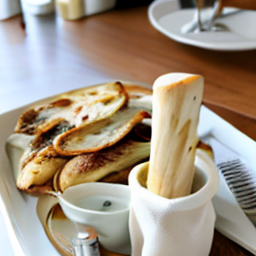}
    & \includegraphics[width=0.15\textwidth]{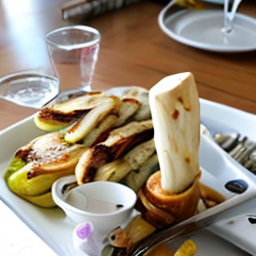} \\
    
    \end{tabular}
    \vspace{-.5em}
    \caption{\textbf{Compositional generation.} \ourmethod enables the compositional image generation of real-world complex scenes. Here, the masked object \emph{(in red)} is the slot to be added (Added item) to the original image resulting in the edited image.}
    \label{fig:appendix:comp_gen_add}
    \vspace{-1em}
\end{figure*}

%% file: tables/nips_tabs/ablation_num_slots.tex
\begin{table}[tbp]
    \centering
    \scriptsize
    \smallskip    
    \begin{tabularx}{\columnwidth}{@{}
    >{\centering\arraybackslash}X
    >{\centering\arraybackslash}X
    >{\centering\arraybackslash}X@{}}
    \toprule
         Num. of slots & mIoU$_{i}$ & mBO$_{i}$\\
         \midrule
         \rowcolor{gray!15} 7 & \textbf{38.9} & \textbf{40.6} \\
         14 & 27.0 & 28.0\\
         21 & 24.2 & 25.3\\
    \bottomrule
    \end{tabularx}
    \vspace{-0.5em}
    \caption{\textbf{Effect of number of slots on \ourmethod.} As with all slot-attention methods, \ourmethod is also sensitive to the number of slots used in the slot-attention module.}
    \label{tab:appendix:effect_num_slots}
    \vspace{-0.5em}
\end{table}

%% file: tables/nips_tabs/ablation_captions.tex
\begin{table}[tbp]
    \centering
    \scriptsize
    \smallskip    
    \begin{tabularx}{\columnwidth}{@{}
    >{\centering\arraybackslash}l
    >{\centering\arraybackslash}X
    >{\centering\arraybackslash}X@{}}
    \toprule
         Caption type & mIoU$_{i}$ & mBO$_{i}$\\
         \midrule
         \rowcolor{gray!15} BLIP-2 & 38.9 & 40.6 \\
         ShareGPT-4V &  38.3 & 40.7\\
         Template-based & \textbf{40.0} & \textbf{41.2}\\
    \bottomrule
    \end{tabularx}
    \vspace{-0.5em}
    \caption{\textbf{Effect of caption generation method on \ourmethod.} We find that our method is robust to the choice of language module to generate the captions. Interestingly, if we can access a ground-truth object occurrence statistics dataset, a template-based caption scheme outperforms learnt language-based methods.}
    \label{tab:appendix:effect_caption}
    \vspace{-0.5em}
\end{table}